\documentclass[ASNA,twocolumn]{USG} \usepackage{anyfontsize} %
\usepackage{bm}

\usepackage{colortbl}   
\usepackage{xcolor}     

\usepackage{steinmetz}

\articletype{RESEARCH ARTICLE}%

\journal{}
\volume{0}
\copyyear{2026}
\startpage{1}

\begin{document}
\title{Adaptive Shared Control with Online Bounded-Rational Human Behavior Estimation}

\transtitle{Adaptive Shared Control with Online Bounded-Rational Human Behavior Estimation}

\author[1]{Henry Ascencio Trejo}
\author[2]{Roel Pieters}
\author[3]{Gokhan Alcan}

\titlemark{Adaptive Shared Control with Online Bounded-Rational Human Behavior Estimation}

\address[]{\orgdiv{Faculty of Engineering and Natural Sciences, }\orgname{Tampere University, }%
\orgaddress{\state{Tampere, }\country{Finland}}}

\corres{ Gokhan Alcan (\email{gokhan.alcan@tuni.fi}) }

\fundingInfo{This work was supported by the Business Finland project AURORA "Automated and Connected Machines" and by the NVIDIA Academic Grant Program through the provision of RTX PRO 6000 Blackwell Max-Q GPUs.}


\transkeywords{shared-control | bounded rationality | approximate dynamic programming | level-k | probabilistic human model}

\abstract[ABSTRACT]{
This work considers adaptive shared human-robot control for nonlinear control-affine systems, where the assumption of a fully rational human is relaxed and the robot adapts its assistance to observed boundedly rational human behavior. 
We use a level-k bounded-rationality model of the two-player game to construct a finite bank of candidate human and robot policies through alternating best-response computations, with the associated value functions and policies approximated using adaptive dynamic programming.
During the shared-control interaction, state-transition residuals compare the measured system evolution with the trajectories predicted by the candidate human policies. 
The residuals are accumulated using a forgetting factor and mapped to a probabilistic human-behavior model over the finite candidate bank.
Rather than selecting a single candidate or averaging stored robot policies, the robot computes a distribution-aware one-step best response by minimizing an expected cooperative cost over the complete estimated human behavior distribution. 
For a quadratic terminal-value approximation and Euler state propagation, this response admits a closed-form solution expressed in terms of the expected human input. 
The proposed methods are evaluated in simulations of a benchmark nonlinear system stabilization task, and of a planar manipulator shared control setup.
The reported results show decreasing Kullback-Leibler divergence between the estimated and simulated human behavior distributions, and a lower accumulated running cost for the robot agent over the shared control interaction period, than the maximum-probability and probability-weighted alternative policies baseline.
}
\maketitle

\section{Introduction}\label{intro}

Robots have been increasingly used to augment human capabilities in tasks involving physical, cognitive, or operational limitations. 
Although fully autonomous systems can be effective in some applications, there are other scenarios where human intervention is necessary, desirable, or required by task, safety, or accountability constraints \cite{abbink2018topology}.
For these scenarios, robots can assist the human-in-the-loop to overcome task-related limitations and improve the performance of the coupled human-robot system \cite{eraslan2020shared}.
Robot assistance can be introduced into the task to complement the human contribution and support the achievement of the cooperative objective \cite{losey2018review, dragan2013policy}. 

Shared-control systems are designed to accomplish joint objectives while considering the interaction between the human and robotic agents involved \cite{abbink2012haptic}. 
In this setting, control authority is not assigned entirely to either the human or the robot. 
Instead, the closed-loop behavior results from the combined effects of the human command, robot assistance, task dynamics, and information available to both agents. 
This balance is especially important in tasks where excessive automation can reduce human agency, while insufficient assistance can leave the human unable to satisfy performance or safety requirements.
A shared control system may account for human behavior when determining the robot's assistive action \cite{zhang2024human}, and adapt the robot assistance to improve joint performance \cite{li2023classification, marcano2020review}.
Robots can use measurements of human behavior to determine an appropriate assistive contribution and improve coordination with the human agent, for example, by prioritizing the human command and introducing only the assistance required by the task \cite{yu2003adaptive, gopinath2016human}. 
Consequently, the human model becomes a control-relevant component of the shared-control loop, rather than only a descriptive representation of the operator.
If this model does not reflect how the human actually reasons about the task and the robot, the assistance may be mistimed, too strong, or inconsistent with the cooperative objective.
More generally, a mismatch between the assumed and observed human behavior may degrade task performance, human authority, and human-robot
coordination.
However, building a model of human behavior that accounts for bounded rationality, and can be leveraged to adapt the robot's assistance during a shared-control interaction remains challenging.
In particular, the robot must infer which candidate bounded-rational behaviors are consistent with the observed system evolution and determine its assistance without reducing this uncertainty to a single assumed human policy.

\begin{figure}[t]
    \centering
    \includegraphics[width=\linewidth]{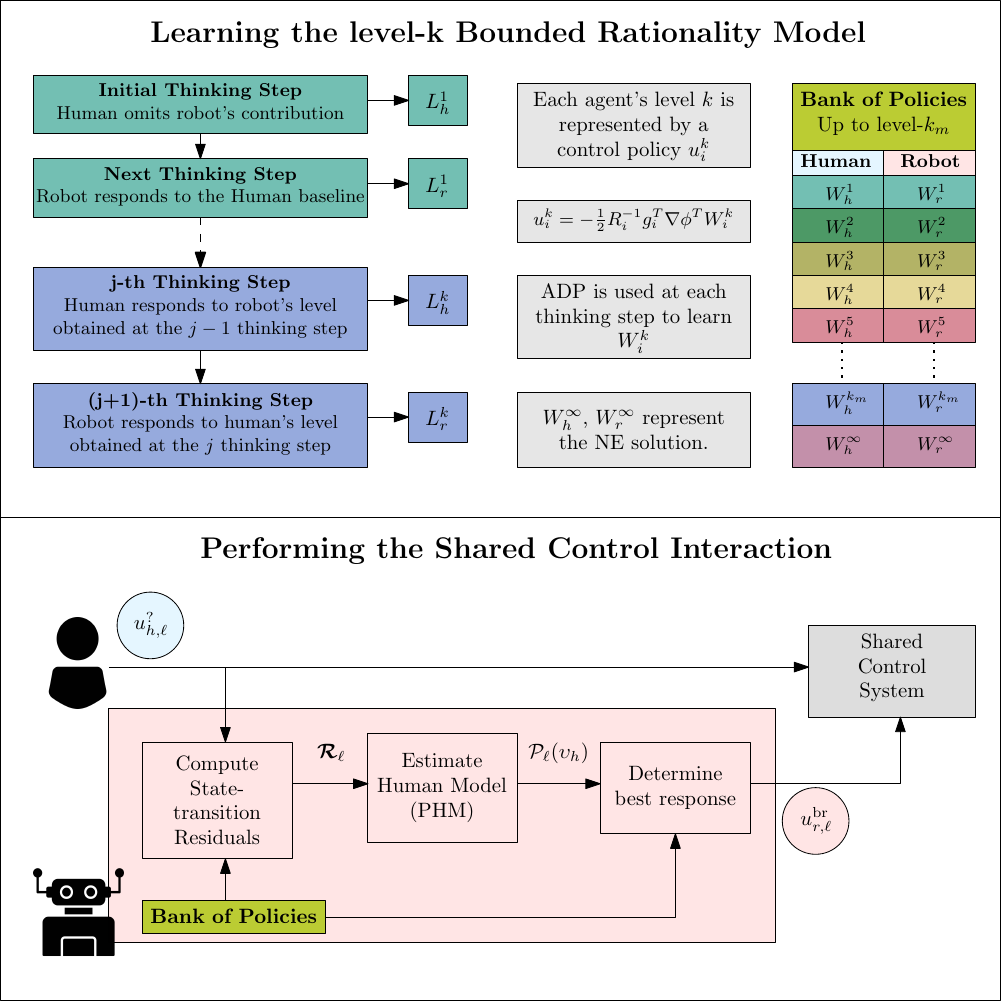}
    \caption{Adaptive Shared Control diagram. An overview of the procedures and mechanisms involved, from learning the level-k model for different bounded rational behaviors, to the shared-control interaction leveraging the human model estimation, and robot's best response.}
    \label{fig:overall_sys}
\end{figure}

\subsection{Related Works}

In the literature, the interaction between a human and a robot agent has been modeled as a two-player differential game. Franceschi et al. \cite{franceschi2023human} explores the interaction over the control of a manipulator system, and Tong et al. \cite{tong2024differential} apply their human-robot cooperative controller to a differential drive system.
In these two-player differential games the players may have aligned objectives and cooperate to accomplish the task.
Non-zero-sum games are useful to represent these types of situations, where both players want a positive outcome without requiring a loss for the other agent \cite{bacsar1998dynamic}.
A solution concept for these games is the Nash equilibrium, in which each agent's policy is a best response to the other and no agent can improve its own objective through a unilateral policy change \cite{li2016framework, vamvoudakis2011multi}. This solution relies on mutually consistent assumptions about the game and the other agent's policy.

When these games involve humans, the fully rational assumption might be too optimistic, may fail to represent observed behavior, and can lead to inconsistent beliefs about the other player \cite{vamvoudakis2022nonequilibrium, costa2006cognition, stahl1995players}.
Modeling human decision-making with bounded rationality is an alternative for representing decision-making under cognitive and informational limitations \cite{dudek2025characterization, li2017game}.
Cognitive hierarchy and level-k models propose representing the non-equilibrium behavior of the players as strategic responses to limited models of the other players and the game \cite{kanellopoulos2019non, abuzainab2016cognitive, camerer2004cognitive}.
In particular, player rationality can be defined by a finite sequence of reasoning steps.
At each level, the player assumes the other players are using a lower level of rationality, producing a strategy that is the best response to the player's current belief. 
The resulting set of policies provides a structured collection of candidate behaviors rather than a single fully rational policy.
Likewise, a widely adopted option for modeling the inherent uncertainty associated with human decision-making involves the use of the softmax function \cite{geng2022human, reverdy2015parameter}.

Adaptive Dynamic Programming (ADP) has proven effective for approximating Nash and non-equilibrium solutions to differential games, as well as for constructing cognitive hierarchy models. Yang et al. \cite{yang2020safe} uses ADP to find the Nash equilibrium solution for a benchmark nonlinear system. Kokolakis et al. \cite{kokolakis2022safety} shows simulations results of solving a pursuit-evasion game for bounded rational UAV systems.
These solutions may leverage online learning techniques using data obtained during the agents' interaction with each other and the environment. Modares et al. \cite{modares2014integral}, and Kokolakis et al. \cite{kokolakis2023bounded} use historical data to learn the solution of the differential games and simulate their methods on benchmark systems, and Dubins vehicle pursuit-evasion problems respectively. 
Furthermore, human-robot interactions have benefited from using these techniques to enable collaborative and optimal solutions for their shared control task by designing assistive controllers. Franceschi et al \cite{franceschi2024design} designs a controller for a robotic manipulator assisting on collaborative transportation. Pezeshki et al. \cite{pezeshki2023cooperative} also uses an ADP strategy on a manipulator system for a rehabilitation task.
In our work, ADP is used before the shared-control interaction to construct banks of level-k human and robot policies, which subsequently serve as the candidate behavior models for online estimation and assistance.

Recent works establish human behavior in a shared control interaction based on a probabilistic model of different bounded rationality levels, which the robot can utilize to determine its assistance on the system. 
Tan et al. \cite{tan2025human} propose modeling human behavior using a probabilistic distribution of the different level-k behaviors found in the training procedure, being able to simulate human input during the shared control interaction. 
They regulate the robot's contribution based on the confidence of human rationality obtained from the pre-computed model. They apply their work on a nonlinear benchmark system, and compute the level-k model for a human cooperating with a robot to control a quadrotor system.
In contrast, the present work proposes a method that continuously estimates the probabilistic human model during the shared control interaction based on the most recent observations of the system's state.
Our aim is to provide a description of the observed human behavior in terms of the available level-k candidates, rather than using those to simulate human policy.
Also, Wu et al. \cite{wu2024human} proposes a similar procedure, where a level-k model contains bounded rational behaviors for the agents in the shared control scenario.
By obtaining and updating a model estimating the rationality level of the human during the interactions, the authors then select the robot response by probabilistically switching to the complementary policy of the most likely human level of rationality. Their experiments simulate a cooperative shared control driver assistance system to evaluate their method.
Alternatively, we define the robot's actions as a response to the full estimated distribution of the human behavior.

In this study, we estimate a model for human behavior based on the bounded rationality framework, and update it continuously during the shared human-robot control interaction to obtain a description of how the human is behaving in the task. 
We leverage that probabilistic model to adapt the robot control input to the system. 
Prior to the interaction, the rationality levels explaining human behavior in the game are computed using ADP to later compare them against the observed human contribution.
The observed state transitions are compared with those predicted by the candidate level-k human policies, and the resulting estimated distribution is used in a one-step optimization problem to compute the robot's next assistance action.
The framework is evaluated in simulation for a stabilization task on a nonlinear benchmark system, and a two-dimensional manipulator human-robot setup using human-model estimation error, accumulated task cost, and agent control effort.

\subsection{Contributions}

The main contributions of this paper are summarized as follows:
\begin{enumerate}
    \item An online method for estimating human behavior in a shared human-robot control interaction based on the level-k bounded rationality framework. By comparing the observed state evolution with the state transitions predicted by the candidate level-k human policies, a probability distribution over the candidate human behaviors is estimated. The human-behavior model is updated throughout the interaction using accumulated state-transition residuals and a forgetting factor.
    
    \item A distribution-aware one-step best response for adaptive robot assistance in shared human-robot control. Rather than selecting the robot policy associated with the most likely human behavior or directly averaging the stored robot policies, the proposed method incorporates the complete estimated human-behavior distribution into a one-step optimization problem to determine the robot control action.

    \item A closed-form realization of the distribution-aware robot response for the quadratic terminal-value approximation and Euler state propagation considered in this work. Under these conditions, the robot action depends on the expected human input and avoids iterative optimization over the candidate behaviors at each interaction interval. The proposed estimator and robot response are evaluated in simulation on a nonlinear stabilization task against maximum-probability and probability-weighted policy baselines.
\end{enumerate}

\section{Problem Formulation}\label{problem}

\subsection{Human-Robot Cooperative System} \label{sec:hr-coop-sys}

We consider a nonlinear control-affine continuous-time system,
\begin{equation} \label{eq:system}
    \dot{x}(t) = f(x(t)) + g_h(x(t)) \: u_h(t) + g_r(x(t)) \: u_r(t) ,\qquad x(0)=x_0
\end{equation}
where $x(t) \in \mathbb{R}^n$ is the system state, $u_{h}(t)\in \mathbb{R}^{m_h}$ is the control input from the human, and $u_r(t)\in \mathbb{R}^{m_r}$ is the control input from the robot. $f(x)$ denotes the drift dynamics of the system, and $g_i(x), \; i \in \{h,r\}$ are the input matrices, with $f:\mathbb{R}^n\rightarrow\mathbb{R}^n$ and $g_i:\mathbb{R}^n\rightarrow\mathbb{R}^{n\times m_i}$.

We assume that $f$ and $g_i$, $i\in\{h,r\}$, are locally Lipschitz on a domain $\Omega\subseteq\mathbb{R}^n$ containing the origin, and that $f(0)=0$. 
The domain $\Omega$ is assumed to be forward invariant under the admissible closed-loop policies.
The admissible policies $\mu_i:\Omega\times\mathbb{R}_{\geq0}\rightarrow\mathbb{R}^{m_i}$ are assumed to be locally Lipschitz in $x$ and piecewise continuous in $t$ and to render the closed-loop system forward complete with finite infinite-horizon cost. 
Consequently, the closed-loop dynamics admit a unique state trajectory for every initial condition $x_0\in\Omega$, and the origin is an equilibrium of the zero-input dynamics.
For policies intended to stabilize the origin, we additionally require $\mu_h(0,t)=\mu_r(0,t)=0$.

In this game, the cooperative running cost $r_{c}$ and the associated human and robot performance indices  $J_i$, $i\in\{h,r\}$ are defined as
\begin{equation} \label{eq:performance_cost}
    J_i(x_0, u_h, u_r) = \int_0^\infty r_{c}\bigl(x(\tau), u_h(\tau), u_r(\tau)\bigr)\,d\tau
\end{equation}
with 
\begin{equation} \label{eq:coop_reward}
    r_{c}(x, u_h, u_r) = M(x) + \sum_{j \in \{h, r\}}u_j^\top  R_{j} u_j
\end{equation}
We assume that $M:\Omega\rightarrow\mathbb{R}_{\geq0}$ is continuous and positive definite with $M(0)=0$, and that $R_i=R_i^\top\succ0$ for $i\in\{h,r\}$.

The Nash equilibrium (NE) is defined as a pair of admissible policies ($u_h^\star$, $u_r^\star$) satisfying
\begin{align} \label{eq:nash_equi_h}
    J_h(x_0, u_h^\star, u_r^\star) \leq J_h(x_0, u_h, u_r^\star), \quad \forall u_h \in\mathcal{U}_h
    \\
    J_r(x_0,u_h^\star, u_r^\star) \leq J_r(x_0, u_h^\star, u_r), \quad \forall u_r \in\mathcal{U}_r\label{eq:nash_equi_r}
\end{align}
where $\mathcal{U}_h$ and $\mathcal{U}_r$ denote the admissible human and robot policy sets, respectively.

\subsection{Shared Human-Robot Control} \label{sec:shared_control}

The shared human-robot control system in \eqref{eq:system} permits the human and robot to contribute
simultaneously to task execution.
While the dynamics specify how the two control inputs affect the state, the policies generating these inputs depend on the agents' decision-making processes.
A fully rational human is represented by the equilibrium policy $u_h^\star$ under a mutually consistent model of the game and of the robot policy $u_r^\star$.
However, in real-life scenarios, the human operator may form an incomplete or inaccurate model of the robot policy \cite{camerer2004cognitive}, due to limited experience, limited information or differences between human and robot computational capabilities \cite{kyriakos2020synchronous}.

During the shared-control interaction, the human policy is not assumed to coincide with $u_h^\star$. Instead, let the human input be generated by an unknown, possibly time-varying admissible policy
\begin{equation}
    u_h(t)=\mu_h\bigl(x(t),t\bigr),
    \qquad \mu_h\in\mathcal{U}_h,
\end{equation}
whose behavior may reflect bounded reasoning, uncertainty, or changing assumptions about the robot.
The robot does not know $\mu_h$ a priori and must determine its assistance causally from the information available during the interaction. Let $\mathcal{I}_t$ denote the available history of measurements and applied robot inputs up to time $t$. A general adaptive shared-control policy can then be written as
\begin{equation}
    u_r(t)=\mu_r\bigl(x(t),\eta(t)\bigr),
    \qquad
    \eta(t)=\mathcal{E}(\mathcal{I}_t),
\end{equation}
where $\eta(t)$ is an online description of the observed human behavior and $\mathcal{E}$ is an estimation or inference mechanism. 
\\ 
\par \textbf{Adaptive shared-control problem.}
Given the nonlinear system \eqref{eq:system}, the cooperative performance index \eqref{eq:performance_cost}, the admissible policy sets $\mathcal{U}_h$ and $\mathcal{U}_r$, and the information history $\mathcal{I}_t$, determine an online behavior-description mechanism $\mathcal{E}$ and a causal robot-assistance policy $\mu_r$ such that:
\begin{enumerate}[(i)]
    \item $\eta(t)=\mathcal{E}(\mathcal{I}_t)$ is updated during the interaction and captures the behaviorally relevant uncertainty associated with the unknown human policy,
    \item the robot action adapts to the current state and the available human-behavior information rather than relying on the fixed assumption $u_h=u_h^\star$,
    \item the resulting human--robot closed loop remains admissible and pursues the cooperative task while seeking to reduce the performance index in \eqref{eq:performance_cost}.
\end{enumerate}
For the stabilization problem considered in this paper, the desired task behavior is convergence of the state toward the origin with finite control effort.

\section{Adaptive Shared Control Framework} \label{asc_sys_fw}

We propose a shared control framework that adapts the robot's contribution to the system based on the estimated distribution over candidate boundedly rational human behaviors.
As presented in Figure \ref{fig:overall_sys}, we establish two stages for providing adaptive robot assistance in a shared-control task.

The first stage focuses on \textbf{learning the candidate level-k behaviors} of the human and robot agents in the two-player game (Section \ref{sec:learning_lk_br}). 
We follow the bounded rationality and ADP constructions of \cite{tan2025human, kokolakis2023bounded} to iteratively construct candidate
policies through a finite number of thinking steps.
As a result, we obtain a \textit{bank of policies} storing the corresponding critic weights and feedback policies associated with the retained level-k candidates.

The second stage determines the adaptive robot assistance while \textbf{performing the shared-control interaction} (Section \ref{sec:performing_sc_int}).
We consider a human whose policy is unknown to the robot.
Rather than assuming that the human applies the full-rationality equilibrium policy, the robot describes the observed behavior relative to the previously learned candidate human policies.
During the interaction, a probability distribution is updated from state-transition residuals that compare the measured system evolution with the evolution predicted under each candidate level-k behavior.
The resulting probabilistic human model (PHM) describes the observed behavior relative to the candidate policy bank
It therefore represents uncertainty over the available behavior models but does not identify arbitrary human behavior outside the policy bank.
Finally, the robot uses the complete estimated distribution in a one-step optimization problem to compute a distribution-aware assistance action.

\section{Construction of the Candidate Level-k Policy Bank} \label{sec:learning_lk_br}

This section first introduces the full-rationality solution of the cooperative game, then defines a finite bounded-rationality recursion, and finally describes the ADP mechanism used to approximate and store the candidate policies. The Nash-equilibrium solution provides a full-rationality or infinite-level reference and may also be used to approximate the terminal value function. The online PHM only requires a finite bank of distinguishable candidate policies.

\subsection{Solution of the Cooperative Game} \label{sec:sol_coop_game}

The Nash-equilibrium (NE) solution for the system in \eqref{eq:system} is a mutual best response pair ($u_h^\star, u_r^\star$) satisfying the conditions in \eqref{eq:nash_equi_h}--\eqref{eq:nash_equi_r}. For a given admissible policy pair, the value function of agent $i\in\{h,r\}$ is
\begin{equation}
    V_i(x;u_h,u_r)
    =
    \int_t^\infty
    r_c\bigl(x(\tau),u_h(\tau),u_r(\tau)\bigr)\,d\tau,
    \qquad x(t)=x.
\end{equation}

At the NE, each value function satisfies the best-response relation
\begin{equation} \label{eq:optimal_value_fun}
    V_i^\star(x)
    =
    \min_{u_i\in\mathcal U_i}
    \int_t^\infty
    r_c\bigl(x(\tau),u_i(\tau),u_{-i}^\star(\tau)\bigr)\,d\tau,
    \qquad i\in\{h,r\},
\end{equation}
where $-i$ denotes the complementary agent and the control arguments of $r_c$ are assigned to their corresponding human and robot channels. The state evolves according to \eqref{eq:system}.

The Hamiltonian associated with agent $i$ is
\begin{align} \label{eq:hamilton}
    \mathcal{H}_i(x, \nabla V_i, u_h, u_r) &= r_c(x, u_h, u_r) + 
    \\
    & \quad \quad (\nabla V_i)^\top  \left [ f(x) + g_h(x)u_h + g_r(x) u_r \right ] \nonumber.
\end{align}

The stationarity condition $\partial \mathcal H_i/\partial u_i=0$ yields
\begin{align}
        u_h^\star(x) &= -\frac{1}{2}R_h^{-1}(g_h(x))^\top  \nabla V_h^\star(x) \label{eq:ne_h_policy}
        \\
        u_r^\star(x) &= -\frac{1}{2}R_r^{-1}(g_r(x))^\top  \nabla V_r^\star(x), \label{eq:ne_r_policy}
\end{align}
where $\nabla V_i = \frac{\partial V_i}{\partial x}, \quad i = h, \: r$.

Substituting \eqref{eq:ne_h_policy} and \eqref{eq:ne_r_policy} into the equilibrium Hamiltonian condition \eqref{eq:hamilton} gives the following coupled Hamilton-Jacobi-Bellman (HJB) equations:
\begin{align} \label{eq:hjb}
    0 &= (\nabla V_i^\star)^\top  \left ( f(x) - \frac{1}{2} \sum_{j \in \{ h, r\}} g_j(x) R_j^{-1} g_j(x)^\top  \nabla V_j^\star \right ) + 
    \\
    & \quad \quad M(x) + \frac{1}{4} \sum_{j \in \{ h, r\}} (\nabla V_j^\star)^\top  g_j(x) R_j^{-1} g_j(x)^\top  \nabla V_j^\star \nonumber
\end{align}

\subsection{Finite Level-k Bounded Rationality Recursion} \label{level_k}

As introduced in Section \ref{asc_sys_fw} the bounded rationality framework constructs a finite collection of policies by alternating best-response computations.

To construct each level-k policy, the procedure modifies the full-rationality construction in Section \ref{sec:sol_coop_game} by restricting each agent's model of the other agent to a previously computed policy
The successive thinking steps reflect changes in the agents' beliefs about the other player's policy

The level-k modeling starts by defining the initial levels (level-0 and level-1).
Subsequent policies are obtained by alternately computing the best response of one agent to the most recently available policy of the other agent.

\subsubsection{Initial Human Policy and First Robot Response} \label{sec:init_h_policy}

The starting level-0 focuses on the human baseline behavior. The level-0 human is modeled as non-strategic with respect to the robot and therefore computes its policy without including the robot contribution in its internal dynamics model.

The level-0 human solves
\begin{align}
    V_{h}^0(x_0) &= \min_{u_h^0} \int_t^\infty (M(x) + (u_h^0)^\top  R_{h} u_h^0) \space d\tau
    \\
    & \quad \text{s.t.} \quad \dot{x} = f(x) + g_h(x) \: u_h^0
\end{align}

The resulting level-0 human policy is
\begin{equation} \label{human_level_0}
    u^0_h = -\frac{1}{2} R_h^{-1}(g_h(x))^\top  \nabla V_h^0
\end{equation}
where $\nabla V_h^0 = \frac{\partial V_h^0}{\partial x}$, and $V_h^0$ satisfies the following HJB equation
\begin{align}
    0  &= r_c(x, u_h^0) + (\nabla V_h^0)^\top  \left [ f(x) + g_h(x)u_h^0 \right ] 
\end{align}

The subsequent level-1 robot policy is defined as the robot's best response to the human at level-0. Therefore, the robot at the level-1 step assumes interaction with a level-0 human, and solves the optimization problem that minimizes the value function at level-1 $V_r^1$ as
\begin{align}
    V_{r}^1(x_0) &= \min_{u_r^1} \int_t^\infty (M(x) + (u_h^0)^\top  R_{h} u_h^0) + (u_r^1)^\top  R_{r} u_r^1) \space d\tau
    \\
    & \quad \text{s.t.} \quad \dot{x} = f(x) + g_h(x) \: u_h^0 + g_r(x) \: u_r^1
\end{align}

The resulting robot policy at level-1 is then
\begin{equation} \label{robot_level_1}
    u^1_r = -\frac{1}{2} R_r^{-1}(g_r(x))^\top  \nabla V_r^1
\end{equation}
where $\nabla V_r^1 = \frac{\partial V_r^1}{\partial x}$, and $V_r^1$ satisfies the following HJB equation
\begin{align}
    0  &= r_c(x, u_h^0, u_r^1) + (\nabla V_r^1)^\top  \left [ f(x) + g_h(x)u_h^0 + g_r(x)u_r^1 \right ] 
\end{align}

\subsubsection{General Alternating Thinking Step}

After defining the initial levels, an iterative procedure is used to similarly obtain the level-k and level-(k+1) definitions. Following the initial sequence, each step computes the best response to the other agent's previously computed policy.
\\ 
\par \textbf{Level-k.}
The level-k computation step focuses on the human. During this thinking step the human assumes the robot behaves with the immediate lower rationality level (k-1).

The level-k human solves the optimization problem that minimizes the value function at level-k $V_h^k$ as

\begin{align}
    V_{h}^k(x_0) &= \min_{u_h^k} \int_t^\infty (M(x) + (u_h^k)^\top  R_{h} u_h^k) + (u_r^{k-1})^\top  R_{r} u_r^{k-1}) \space d\tau
    \\
    & \quad \text{s.t.} \quad \dot{x} = f(x) + g_h(x) \: u_h^{k} + g_r(x) \: u_r^{k-1}
\end{align}

The resulting human policy at level-k is
\begin{equation} \label{eq:human_level_k}
    u^k_h(x) = -\frac{1}{2} R_h^{-1}(g_h(x))^\top  \nabla V_h^k(x)
\end{equation}

The corresponding HJB equation is

\begin{align}
    0  &= r_c(x, u_h^k, u_r^{k-1}) + (\nabla V_h^k)^\top  \left [ f(x) + g_h(x)u_h^k + g_r(x)u_r^{k-1} \right ] 
\end{align}
\\ 
\par \textbf{Level-(k+1).}
The following thinking step focuses on the robot. A robot at thinking step k+1 assumes interaction with the previously computed level-k human policy.

The level-(k+1) robot solves the optimization problem that minimizes the value function at level-(k+1) $V_r^{k+1}$ as

\begin{align}
    V_{r}^{k+1}(x_0) &= \min_{u_r^{k+1}} \int_t^\infty (M(x(\tau)) + (u_h^k(\tau))^\top  R_{h} u_h^k(\tau)) \\
    & \quad\quad\quad\quad + (u_r^{k+1}(\tau))^\top  R_{r} u_r^{k+1}(\tau)) \space d\tau
    \\
    & \quad \text{s.t.} \quad \dot{x} = f(x) + g_h(x) \: u_h^{k} + g_r(x) \: u_r^{k+1}
\end{align}

The resulting robot policy at level k+1 is then
\begin{equation} \label{eq:robot_level_k_1}
    u^{k+1}_r = -\frac{1}{2} R_r^{-1}(g_r(x))^\top  \nabla V_r^{k+1}
\end{equation}
where $\nabla V_r^{k+1} = \frac{\partial V_r^{k+1}}{\partial x}$, and $V_r^{k+1}$ satisfies the following HJB equation
\begin{align}
    0  &= r_c(x, u_h^k, u_r^{k+1}) + (\nabla V_r^{k+1})^\top  \left [ f(x) + g_h(x)u_h^k + g_r(x)u_r^{k+1} \right ] 
\end{align}

\begin{remark}
    Stability of the policy pair $(u_h^k,u_r^{k+1})$ is established in \cite{tan2025human} under the assumptions stated in that work. Applicability of that result here only requires the present dynamics, cost, policy approximation, and learning conditions to satisfy the cited assumptions.
\end{remark}

\begin{remark}
    The convergence result in \cite{kokolakis2023bounded} may be used to relate the infinite-thinking-step limit to the NE policies, provided that the assumptions of that result hold for the present game. Under those conditions, the limiting policies, denoted by $u_h^\infty$ and $u_r^\infty$, coincide with $u_h^\star$ and $u_r^\star$, respectively.
\end{remark}

\begin{remark} \label{rem:level_indices}
    The indices $k$ and $k+1$ in this subsection denote consecutive thinking steps in the alternating recursion, corresponding to index $j$ in Algorithm \ref{alg:level_k_model}. For online estimation, the retained human candidates are relabeled by the common index set $\mathcal K=\{1,\ldots,k^m\}$ following the human thinking steps $k$. The corresponding robot policies are stored with the human candidates according to the best-response pairing used during construction following the robot thinking steps ${k+1}$.
\end{remark}

\subsection{ADP Approximation of the Candidate Policies}\label{sec:sol_level_k}

To solve the level-k bounded rationality problem, a data-driven ADP learning algorithm is applied offline at each thinking step.

For each agent $i, \: \in \{ h, r \}$, at a particular thinking step $j$, we can approximate the value function $V_i^j$ and its gradient $\nabla V_i^j$ using a critic neural network that is linear in its parameters:
\begin{equation}
    {V}_i^j = (W_i^j)^\top \phi^j(x) + \epsilon_i^j(x)
\end{equation} \label{eq:value_fun_approx}
where $W_i^j\in\mathbb{R}^N$  are the critic weights for the agent $i$, at the thinking step $j$, $\phi(x) \in \mathbb{R}^{N}$ is the critic basis vector, $\epsilon_i(x)$ is the residual errors of the critic network, $N$ is the number of basis functions and neurons in the critic network. The corresponding gradient is
\begin{equation} \label{eq:gradient_aprox_val}
    \nabla V_i^j(x)
    =[\nabla_x\phi^j(x)]^\top W_i^j+\nabla\epsilon_i^j(x),
\end{equation}
where $\nabla_x\phi^j(x)\in\mathbb R^{N\times n}$.

The ideal weights $W_i^j$ are unknown. Their estimates $\hat{W}_i^j$ define
\begin{align}
    \hat V_i^j(x)&=(\hat W_i^j)^\top\phi^j(x),
    \label{eq:est_val}\\
    \nabla\hat V_i^j(x)&=[\nabla_x\phi^j(x)]^\top\hat W_i^j.
    \label{eq:gradient_est_val}
\end{align}

The corresponding approximate policy is
\begin{equation} \label{eq:approx_policies}
    \hat u_i^j(x)
    =-\frac{1}{2}R_i^{-1}g_i(x)^\top
    [\nabla_x\phi^j(x)]^\top\hat W_i^j.
\end{equation}

At thinking step $j$, let $u_{-i}^{j-1}$ denote the fixed policy of the complementary agent. Define the approximate closed-loop vector field
\begin{equation}
F_i^j(x)=f(x)+g_i(x)\hat u_i^j(x)+g_{-i}(x)u_{-i}^{j-1}(x)
\end{equation}
and the regressor
\begin{equation}
\omega_i^j(x)=\nabla_x\phi^j(x)F_i^j(x).
\end{equation}

Substituting \eqref{eq:gradient_est_val} into the Hamiltonian \eqref{eq:hamilton} gives the Hamiltonian residual
\begin{equation}\label{eq:hamilton_approx}
\delta_i^j(x)
=r_c\bigl(x,\hat u_i^j(x),u_{-i}^{j-1}(x)\bigr)
+(\hat W_i^j)^\top\omega_i^j(x).
\end{equation}

The residual $\delta_i^j$ is driven toward zero by the critic update.

A history stack of data of length $Z$ is used together with the instantaneous residual. Let
\begin{align}
    e_{\mathrm{ins},i}^j(t)&=\delta_i^j(x(t)),\\
    e_{\mathrm{his},i}^j(t_\ell,t)
    &=r_c\bigl(x(t_\ell),\hat u_i^j(x(t_\ell)),u_{-i}^{j-1}(x(t_\ell))\bigr)\\
    &+(\hat W_i^j(t))^\top\omega_{\ell,i}^j,
\end{align}
where $\omega_{\ell,i}^j$ is the stored regressor associated with the sample at $t_\ell$. A normalized squared-residual objective is
\begin{equation}
    E_i^j
    =\frac{1}{2}
    \left(
    \frac{(e_{\mathrm{ins},i}^j)^2}{(1+(\omega_i^j)^\top\omega_i^j)^2}
    +\sum_{\ell=1}^{L}
    \frac{(e_{\mathrm{his},i}^j)^2}
    {(1+(\omega_{\ell,i}^j)^\top\omega_{\ell,i}^j)^2}
    \right).
\end{equation}

Using a normalized semi-gradient update gives
\begin{align} \label{eq:learning_policy_k}
    \dot{\hat{W}}_i^j &= -a \frac{\partial E_i^j}{\partial \hat{W}_i^j} \\
    \dot{\hat{W}}_i^j &= -a \left [ \frac{ \omega_i^j \: \left ( e_{\text{ins},\: i}^j \right ) }{(1 + (\omega_i^j)^\top  \omega_i^j)^2} +  \sum_{z=1}^Z  \frac{ \omega_{z, \:i}^j \: \left ( e_{\text{his}, \: i}^j \right ) }{(1 + (\omega_{z, \: i}^j)^\top  \omega^j_{z, \: i})^2}  \right ]
\end{align}
where $a \in \mathbb{R}_{>0}$ is the learning rate determining the speed of convergence for the critic neural network weights.

The history stack
$\bm\omega_i^j=[\omega_{1,i}^j,\ldots,\omega_{Z,i}^j]$
contains samples recorded at $t_1,\ldots,t_Z$. The finite-rank condition
\begin{equation} \label{eq:hist_rank_cond}
\operatorname{rank}(\bm\omega_i^j)=N
\end{equation}
provides excitation over the stored data and is used to relax a persistent-excitation requirement on the instantaneous trajectory.

\begin{remark}
    Uniform ultimate boundedness of the critic-weight error may be invoked from \cite{tan2025human} only when the assumptions of the cited result, including the required rank condition and bounded approximation errors, hold here. The learning rate $a$, initial weights, stopping criterion, and construction of the history stack influence the observed convergence. Related data-driven ADP results also emphasize the need for a sufficiently informative history stack \cite{kokolakis2022safety,vamvoudakis2015asymptotically}.
\end{remark}

\begin{algorithm}
    \caption{Construction of the candidate level-k policy bank}\label{alg:level_k_model}
    \begin{algorithmic}
        \Require $x_0, \: R_h, \:R_r, \:a, \: k^m$
        \Procedure {}{}
        \State Set $W_r^{-1} = \bm{0}$ \Comment{No robot contribution}
        \For {$k=1 \: , \dots \:, k^m$}:
            \For {$i \in \{ h, \: r \}$}:
                \State Set $q = i^c$ \Comment{Complementary agent}
                \If {i = h}
                    \State $j = 2k - 2$ \Comment{$h$ thinking step}
                \Else
                    \State $j = 2k - 1$ \Comment{$r$ thinking step}
                \EndIf
                \State Initialize $W_i = W_{i, 0}$, and $W_q = W_q^{j-1}$
                \Procedure {learn $u_i^{j}$ until convergence}{}
                    \State Propagate \eqref{eq:system} with $u_q^{j-1}$, and $u_i^{j}$
                    \State Update $u_i^j$ using \eqref{eq:approx_policies} and \eqref{eq:learning_policy_k}
                \EndProcedure
                \State Update bank of policies with $W_i^j \leftarrow W_i$
            \EndFor
        \EndFor
        \State Initialize $W_h^\infty = W_{h, 0}$, and $W_r^\infty = W_{r, 0}$
        \Procedure {learn $u_h^{\infty},\: u_r^{\infty}$ until convergence}{}
            \State Propagate \eqref{eq:system} with $u_h^{\infty}$, and $u_r^{\infty}$
            \State Update $u_h^\infty$ and $u_r^\infty$ using \eqref{eq:approx_policies} and \eqref{eq:learning_policy_k}
        \EndProcedure
        \State Update bank of policies with $W_h^{2k^m} \leftarrow W_h^\infty$ \State Update bank of policies with $W_r^{2k^m + 1} \leftarrow W_r^\infty$
    \EndProcedure
    \end{algorithmic}
\end{algorithm}

\section{Online Adaptive Shared-Control Interaction} \label{sec:performing_sc_int}

The online stage uses the learned human policy bank to estimate which candidate behaviors are most consistent with the measured state evolution and then computes the robot assistance from the complete estimated distribution.

Let $t_\ell=\ell T_{\mathrm{int}}$ and define the $\ell$th interaction interval as $\mathcal T_\ell=[t_{\ell-1},t_\ell]$, for $\ell\geq 1$. 
The robot action $u_{r,\ell-1}$ is computed at $t_{\ell-1}$ and held constant over $\mathcal T_\ell$. 
The human input is generated by the unknown policy and need not be measured directly. 
The estimator uses the observed state trajectory, or sampled state measurements, together with the known robot action. 
The candidate behavior identity is assumed to vary slowly enough that a single distribution can describe each interval. 
The candidate feedback action may still vary with the state.

\subsection{State-transition Residuals}

To estimate and update the PHM, the robot must infer information about the observed human behavior. 
We assume that only the system state and the applied robot action are available, and infer the effect of the agents' combined inputs through the state transitions. 
Our goal is to map the observed human behavior to the candidate human level-k behaviors previously obtained in Section \ref{sec:sol_level_k}. 
By comparing the measured state evolution with the evolution predicted under each candidate human policy, we can construct a description of $\upsilon_h$ based on the candidate level-k behaviors.
We compute the state-transition residuals $\rho$ to represent the discrepancy between the measured and candidate-predicted state evolutions.

The level-k model gives the human set of candidate behavior indices $\mathcal{K} = \{ 1, \dots, k^m \}$. For each $j \in \mathcal{K}$, the state trajectory predicted over $\mathcal T_\ell$ is generated from the measured initial state, the known robot action, and the candidate feedback policy $\hat u_h^j$ as
\begin{align}
\dot{\hat x}_{\ell}^{j}(\tau)
&=f\!\left(\hat x_{\ell}^{j}(\tau)\right)
+g_h\!\left(\hat x_{\ell}^{j}(\tau)\right)
\hat u_h^j\!\left(\hat x_{\ell}^{j}(\tau)\right)
+g_r\!\left(\hat x_{\ell}^{j}(\tau)\right)u_{r,\ell-1},
\nonumber\\
\hat x_{\ell}^{j}(t_{\ell-1})
&=x^{\mathrm{obs}}(t_{\ell-1}),
\qquad \tau\in\mathcal T_\ell.
\label{eq:candidate_predicted_trajectory}
\end{align}

Equivalently, the predicted endpoint is

\begin{equation} \label{eq:level_k_behav}
    \hat{x}_{\ell}^{j}(t_\ell)
    = \theta_{T_{\mathrm{int}}}\!\left(
    x^{\mathrm{obs}}(t_{\ell-1}),
    u_{r,\ell-1},
    \hat{u}^j_h
    \right),
\end{equation}
where $\theta_{T_{\mathrm{int}}}$ denotes the closed-loop flow over one interaction interval.

Here, $x^{\mathrm{obs}}(t_{\ell-1})$ is the measured state at the beginning of the interval, $u_{r,\ell-1}$ is the robot action held over $\mathcal T_\ell$, and $\hat x_\ell^j$ is simulated internally using the candidate policy $\hat u_h^j$.

The state-transition residual for candidate $j$ over interval $\mathcal T_\ell$ is
\begin{equation} \label{eq:state_trans_res}
    \rho_\ell^j
    = \int_{t_{\ell-1}}^{t_\ell}
    \left\|x^{\mathrm{obs}}(\tau)-\hat{x}_{\ell}^{j}(\tau)\right\|_2\,d\tau.
\end{equation}

The state-transition residual vector $\bm{\rho}_\ell = [\rho_\ell^1,\rho_\ell^2,\dots,\rho_\ell^{k^m}]^\top$ contains the residuals for each of the available candidate human behaviors in $\mathcal{K}$.

\begin{remark}
    The observed system state $x^{obs}$ corresponds to the measured evolution generated by the unknown human input and the applied robot action. In contrast, $\hat{x}_{\ell}^{j}$ is the internally simulated evolution generated by candidate policy $\hat u_h^j$ and the same robot action. Using the same initial state and robot action isolates the discrepancy associated with the candidate human model, subject to plant-model and measurement errors
\end{remark}

\begin{remark}
    Using state transitions avoids requiring direct measurement of the human input and therefore reduces the sensing requirements. However, measurement noise and plant-model mismatch also contribute to the residuals and may reduce the distinguishability of the candidate behaviors. 
\end{remark}

\begin{figure*}[tb]
    \centerline{\includegraphics{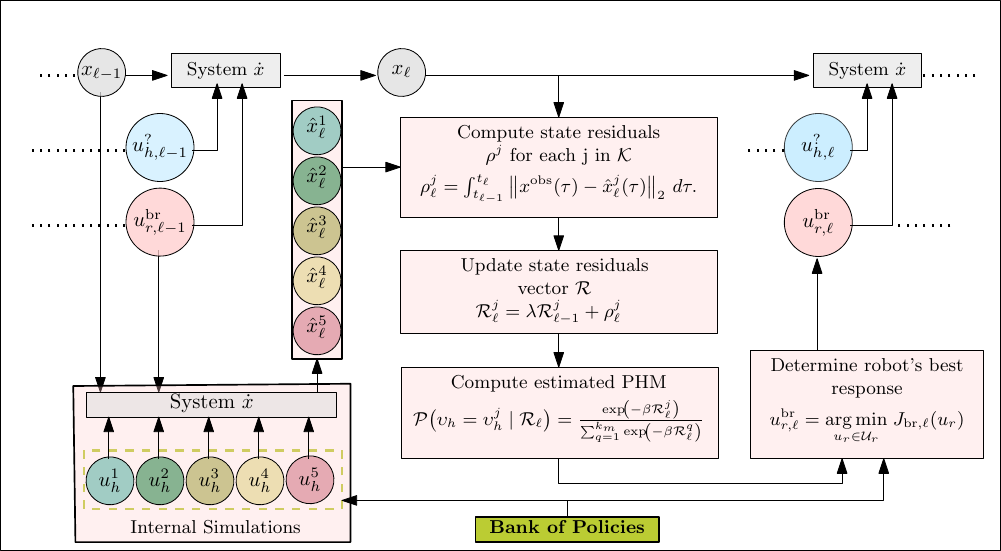}}
    \caption{Proposed methods and processes during an interval of interaction of the shared control system. The human input gets assisted by the robot's best response based on the estimated PHM.}
    \label{fig:sc_interaction_process}
\end{figure*}

\subsection{Probabilistic Human-Behavior Model and Online Update} \label{sec:estimating_phm}

Using the bounded rationality framework and the iterative thinking procedure we obtain the level-k model for the different behaviors of the human and robot agents interacting in the two-player game. The human level-k model, up to a maximum level $k^m$ is represented by a bank of policies $\bm{B}$ where the learned weights $W_h^j,\: j \in \{ 1, \dots, k^m \}$ are stored. The different candidate policies provide a finite model class for describing the observed human behavior. We denote candidate behavior $j$ by $\upsilon_h^j$, corresponding to the feedback policy $\hat u_h^j$

Since the measured transition may not coincide exactly with any single candidate, the PHM assigns a probability to each candidate behavior. A softmax mapping converts smaller transition residuals into larger probabilities. An instantaneous distribution based only on interval $\mathcal T_\ell$ can be defined as
\begin{equation} \label{eq:phm_softmax}
    p_{\ell,\mathrm{inst}}^{(j)}
    =\mathcal P\!\left(\upsilon_h=\upsilon_h^j\mid\bm\rho_\ell\right)
    =\frac{\exp\!\left(-\beta\rho_\ell^j\right)}
    {\sum_{q=1}^{k^m}\exp\!\left(-\beta\rho_\ell^q\right)},
    \qquad j\in\mathcal K,
\end{equation}
where $\beta>0$ is an inverse-scale parameter controlling the concentration of the distribution.

A larger value of $p_{\ell,\mathrm{inst}}^{(j)}$ indicates that candidate $j$ is more consistent with the state transition observed during $\mathcal T_\ell$. This distribution describes behavior only relative to the finite policy bank and does not identify arbitrary behavior outside that model class.

During the shared human-robot control interaction, the robot is aware of the different potential human behaviors $\upsilon_h^j$ represented by the bank of policies. 
At the same time, the actual human policy is unknown to the robot.

\begin{remark}
    The PHM is assumed to be approximately constant within each interaction interval. This does not require the candidate human action to be constant, because each candidate is a state-feedback policy. The robot action is held constant over the interval according to the timing convention above. 
\end{remark}

Rather than selecting a single candidate level, our method maintains a distribution over the entire candidate bank.
Evidence from successive interaction intervals is accumulated using a forgetting factor, and the resulting scores are mapped to the PHM used by the robot-response computation.

As previously mentioned, the process of obtaining the PHM runs continuously during the shared human-robot control interaction.

At time $t_\ell$, after the state data from $\mathcal T_\ell$ have been collected, the residual vector $\bm\rho_\ell$ is computed. To combine the current evidence with previous intervals, we introduce a forgetting factor $\lambda\in[0,1]$ and update the accumulated residual associated with each candidate as

\begin{equation} \label{eq:update_state_trans_vec}
    \mathcal{R}_\ell^j = \lambda \mathcal{R}_{\ell-1}^j + \rho_\ell^j \;, \qquad \mathcal R_0^j=0,\quad j\in\mathcal K.
\end{equation}

where $\mathcal{R}_\ell$ is the accumulated state-transition residual with forgetting factor. 
Similarly, $\bm{\mathcal R}_\ell=[\mathcal R_\ell^1,\dots,\mathcal R_\ell^{km}]^\top$ collects the accumulated residuals. 
Values $\lambda<1$ progressively discount older evidence, whereas $\lambda=1$ retains all past residuals without forgetting.

Following the softmax mapping in \eqref{eq:phm_softmax}, the updated PHM is
\begin{equation} \label{eq:phm_softmax_acc}
    p_\ell^{(j)}
    =\mathcal P\!\left(\upsilon_h=\upsilon_h^j\mid\bm{\mathcal R}_\ell\right)
    =\frac{\exp\!\left(-\beta\mathcal R_\ell^j\right)}
    {\sum_{q=1}^{k^m}\exp\!\left(-\beta\mathcal R_\ell^q\right)},
    \qquad j\in\mathcal K.
\end{equation}

\begin{remark}
    A lower state-transition residual indicates that the candidate policy predicts a state evolution closer to the measured trajectory. The softmax mapping therefore assigns a larger probability to candidates with smaller accumulated residuals.
\end{remark}

\subsection{Distribution-Aware One-Step Best Response} \label{sec:robot_br}

The description of the human behavior $\upsilon_h$ is provided by the PHM distribution over the candidate human policies.
By avoiding selection of a single human candidate during a $T_{int}$, we are interested in computing a robot action that accounts for the complete PHM.

We propose solving a one-step optimization problem against the complete estimated distribution to determine the robot's action for the next $T_{\text{int}}$. 
Using a forward-Euler approximation of the running-cost integral over one interaction interval, we define

\begin{equation} \label{eq:cost_best_response}
    J_{\mathrm{br},\ell}(u_r)
    = \sum_{j\in\mathcal{K}} p_\ell^{(j)}\!\left[
    T_{\mathrm{int}}\,
    r_c\!\left(x_\ell,\hat u_h^j(x_\ell),u_r\right)
    +V_f\!\left(
    \theta_{T_{\mathrm{int}}}\!\left(x_\ell,u_r,\hat u_h^j\right)
    \right)
    \right],
\end{equation}
where $x_\ell=x^{\mathrm{obs}}(t_\ell)$ and $V_f$ is a terminal-value approximation.

Then, the distribution-aware one-step best response is defined by
\begin{equation} \label{eq:robot_best_resp}
    u_{r,\ell}^{\mathrm{br}}
    =\arg\min_{u_r\in\mathcal U_r}
    J_{\mathrm{br},\ell}(u_r). 
\end{equation}

where $u_{r}^{\text{br}}$ is the robot's distribution-aware one-step response, $p_\ell^{(j)} = \mathcal{P}(\upsilon_h = \upsilon_h^j)$ is the probability assigned to candidate human behavior $j$, $r_c$ is the cooperative running-cost function defined in the problem formulation, $V_f$ is an approximate terminal cost function obtained from a selected learned value-function approximation, and $\theta$ is a flow map obtained by propagating the system over $T_{\text{int}}$. 
Solving \eqref{eq:robot_best_resp} returns the adaptive robot response that minimizes the specified approximate one-step expected cost.

\subsection{Closed-Form Response for a Quadratic Terminal Cost} \label{sec:br_response_cfsol}

When considering a state vector $\bm{X} \in \mathbb{R}^{n}$, a quadratic terminal-value approximation for $V_f$, and an Euler discretization of $\theta$, the problem \eqref{eq:robot_best_resp} admits a closed-form solution when the resulting quadratic objective is strictly convex in $u_r$.

First, given the quadratic basis for $\phi$ and state $\bm{X}$, the terminal-value approximation can be written as
\begin{equation}
    {V}_f(\bm X) = \bm{X}^\top P\bm{X},\qquad P=P^\top.
\end{equation}

For the state containing consecutive related variable pairs, and quadratic basis, the symmetric matrix is
\begin{equation}\label{eq:Pmat}
    P = \begin{bmatrix}
    w_1 & w_2/2 & \cdots & 0 & 0 \\
    w_2/2 & w_3 & \cdots & 0 & 0 \\
    \vdots & \vdots & \ddots & 0 & 0 \\
    0 & 0 & \cdots & w_{z-2} & w_{z-1}/2 \\
    0 & 0 & \cdots & w_{z-1}/2 & w_z
    \end{bmatrix},
\end{equation}
where $z = 3n/2$, $n \geq 2$, and $P \succeq 0$, and is composed of the learned weights $\hat{W}^\star = [w_1,\dots,w_z]^\top$ associated with the selected terminal-value approximation.

Applying an Euler discretization to $\theta$ over $T_{\mathrm{int}}$ we obtain
\begin{equation}
    \begin{aligned}
        \theta_{T_{\mathrm{int}}}\!\left(\bm X_\ell,u_r,\hat u_h^j\right)
    &\approx  \bm X_\ell \\ 
    & +T_{\mathrm{int}}\!\left[
    f(\bm X_\ell)
    +g_h(\bm X_\ell)\hat u_h^j(\bm X_\ell)
    +g_r(\bm X_\ell)u_r
    \right].        
    \end{aligned}
\end{equation}

Define 
\begin{align}
    G_r &= T_{\mathrm{int}}\,g_r(\bm{X}_\ell) \; , \\
    \bm{d}^{(j)} &= \bm{X}_\ell + T_{\mathrm{int}} \bigl[f(\bm{X}_\ell) + g_h(\bm{X}_\ell)\hat{u}_h^j(\bm{X}_\ell) \bigr].    
\end{align}

Then
\begin{equation}
    V_f\!\left(\theta_{T_{\mathrm{int}}}(\bm{X}_\ell, u_r, \hat{u}_h^j)\right) = \bigl( \bm{d}^{(j)} + G_r\,u_r \bigr)^\top P ( \bm{d}^{(j)} + G_r\,u_r \bigr).
\end{equation}

Rewriting the objective in \eqref{eq:cost_best_response} we have
\begin{equation}
    \begin{aligned}
        J_{\mathrm{br},\ell}(u_r)
        &=  \bm{X}^\top Q \bm{X} + \sum_{j \in \{h, r\}}u_j^\top  R_{j} u_j \\
        &+ \bigl( \bm{d}^{(j)} + G_r\,u_r \bigr)^\top P ( \bm{d}^{(j)} + G_r\,u_r \bigr).
    \end{aligned}
\end{equation}

After dropping the terms constant in $u_r$, the objective becomes

\begin{equation}
    \begin{aligned}
        J_{\mathrm{br},\ell}(u_r)
        &=u_r^\top\!\left(T_{\mathrm{int}}R_r+G_r^\top P G_r\right)u_r \\
        &+2\!\left(\sum_{j=1}^{k^m}p_\ell^{(j)}\bm d^{(j)}\right)^\top
        P G_r u_r \\
        &+\mathrm{const}.        
    \end{aligned}
\end{equation}

We set $\frac{\partial J_{\text{br}}}{\partial u_r} = 0$ to get
\begin{equation}
    \bm{u}_{r,\ell}^{\mathrm{br}}
    =-\left(T_{\mathrm{int}}R_r+G_r^\top P G_r\right)^{-1}
    G_r^\top P\,\bar{\bm d}_\ell,
\end{equation}
provided that $T_{\mathrm{int}}R_r+G_r^\top P G_r\succ0$ and
\begin{align}
    &\bar{\bm{d}}_\ell = \bm{X}_\ell + T_{\mathrm{int}} \bigl[ f(\bm{X}_\ell) + g_h(\bm{X}_\ell) \bar{\bm{u}}_{h,\ell} \bigr] \\
    &\bar{\bm{u}}_{h,\ell} = \sum_{j=1}^{k^m} p_\ell^{(j)} \hat{u}_h^j(\bm{X}_\ell).    
\end{align}

\begin{remark}
    The component $\bar{\bm{u}}_{h,\ell}$ appearing in the closed-form solution is the expected human action, summarizing the full human model distribution. This solution reduces the computational burden of the robot-response calculation by avoiding numerical optimization of \eqref{eq:robot_best_resp} at every interaction interval. Candidate forward simulations are still required to compute the state-transition residuals used by the PHM.
\end{remark}

\begin{algorithm}
    \caption{Performing the Online Shared Control Interaction} \label{alg:sc_inter}
    \begin{algorithmic}
        \Require $x_0, \:R_r, \:T_{\mathrm{int}}, \:\bm{B}, \:\lambda, $
        \Procedure {}{}
        \State Initialize $\mathcal{P}(\upsilon_h)$, and $u_r^{\mathrm{br}}$ 
        \For {$\ell = 1, \cdots, \mathcal{L}$}
            \State Propagate \eqref{eq:system} with $u_h^?$, and $u_r^{br}$ over [$(\ell - 1) T_{\mathrm{int}}$, $\ell T_{\mathrm{int}}$]
            \State Take state observation $x^{\mathrm{obs}}$ at $\ell  T_{\mathrm{int}}$
            \State Simulate candidate level-k human behaviors as in \eqref{eq:level_k_behav} over [$(\ell - 1) T_{\mathrm{int}}$, $\ell T_{\mathrm{int}}$]
            \State Compute state-transition residuals $\bm{\rho}$ using \eqref{eq:state_trans_res}
            \State Update state-transition residuals vector $\bm{\mathcal R}$ using \eqref{eq:update_state_trans_vec}
            \State Estimate human behavior via \eqref{eq:phm_softmax_acc} and update $\mathcal{P}(\upsilon_h)$
            \State Solve \eqref{eq:robot_best_resp} to obtain $u_r^{\mathrm{br}}$ and apply it at next interval
        \EndFor
        \EndProcedure
    \end{algorithmic}
\end{algorithm}

\section{Simulation Experiments}\label{sim:overview}

To evaluate our proposed methods, we present simulation experiments involving a shared human-robot control interaction for two different systems. First, we focus on applying our proposed method in a benchmark nonlinear system. Then, we use a planar robotic manipulator system jointly controlled by a simulated human and a robot. In each of these contexts, we address the following questions in our procedures:
\begin{itemize}
    \item Does the probabilistic estimation method of Section \ref{sec:estimating_phm} accurately describe the observed human behavior relative to the finite level-$k$ candidate bank?
    \item Does the distribution-aware one-step response in Section \ref{sec:robot_br} reduce the accumulated running cost, compared with the two baselines that also use the estimated PHM?
\end{itemize}

\subsection{Benchmark Nonlinear System}

\begin{figure*}[b]
    \centerline{\includegraphics{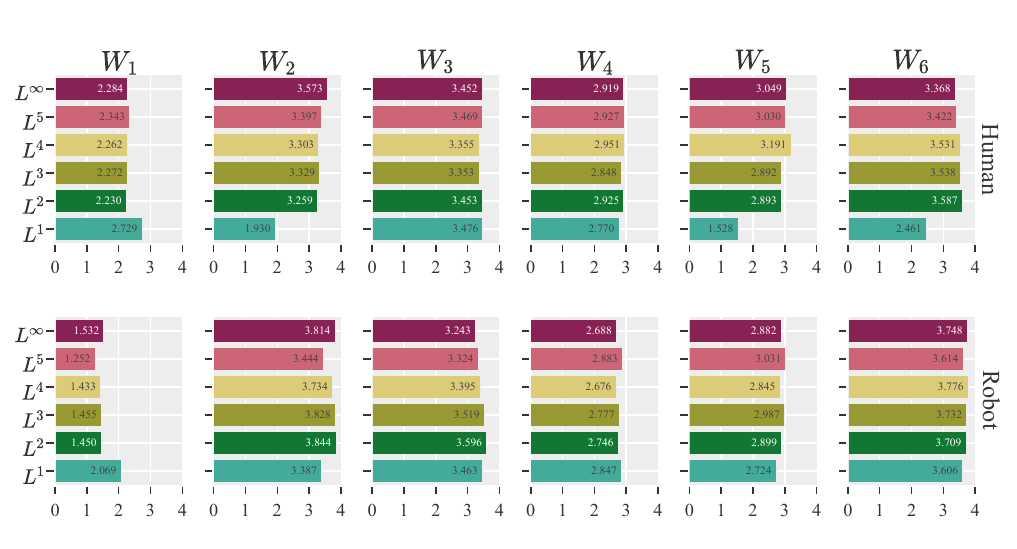}}
    \caption{Final weights of each human (top) and robot (bottom) levels of rationality (including the infinite level) acquired during the learning procedure. Each column corresponds to one of the weights of a specific layer of the critic neural network. These stored weights represent the bank of policies available for different level-k behaviors.}
    \label{fig:level-k_full}
\end{figure*}

\subsubsection{System Description}

To evaluate the performance of our approach we use a nonlinear control-affine dynamical system with separate human and robot input channels based on the experiment designs from \cite{vamvoudakis2011multi, yang2020safe, tan2025human}, and extend the original system's state dimension by duplicating the state derivatives $\dot{x}_1, \dot{x}_2$, to define $\dot{x}_3, \dot{x}_4$ respectively:
\begin{align} \label{system_experiment}
    \dot{x}_1 &= x_2
    \\
    \dot{x}_2 &= -x_2 -\frac{1}{2}x_1 +\frac{x_2(\cos(2x_1)+2)^2}{4} +\frac{x_2(\sin(2x_1)+2)^2}{4} \nonumber \\ & \quad + (\cos(2x_1) + 2) u_1^h + (\sin(4x_1^2) + 2) u_1^r  \nonumber
    \\ 
    \dot{x}_3 &= x_4 \nonumber
    \\
    \dot{x}_4 &= -x_4 -\frac{1}{2}x_3 +\frac{x_4(\cos(2x_3)+2)^2}{4} +\frac{x_4(\sin(2x_3)+2)^2}{4} \nonumber \\ & \quad + (\cos(2x_3) + 2) u_2^h + (\sin(4x_3^2) + 2) u_2^r \nonumber
\end{align}

where $x \in \mathbb{R}^4$ is the system state, and $u^i \in \mathbb{R}^2, i= {h, r}$ are the control inputs of each agent (human and robot respectively).

The shared control task is to stabilize the system's state $x$, so that it converges to the origin from an arbitrary initial condition. During the shared control interaction, both human and robot apply their control inputs to the system by following the process in Figure \ref{fig:sc_interaction_process}. Prior to this, the rationality levels describing the potential behavior of the players during the game are obtained using the methods in Section \ref{sec:learning_lk_br} and become available to later use through the bank of policies.

\subsubsection{Learning the rationality levels model}

To obtain the level-k model representing the potential behaviors of the human and robot in the game, the procedure explained in Algorithm \ref{alg:level_k_model} is used. Two single critic neural networks (NNs) approximate each agent value function as in Section \ref{sec:sol_level_k} while acquiring each level of rationality. For each agent $i$, for $i\in\{h,r\}$, we compute each rationality level $j$, for $j \in \{1, \dots, k^m\}$ with $k^m$ being the maximum level to acquire. The total thinking steps is $2k^m$. For the first thinking step that acquires the initial human policy described in Section \ref{sec:init_h_policy}, the human's agent NN initial weights are $W_{h,0}^1 = [4 \; 4 \; 4 \; 4 \; 4 \; 4]^\top + w^1$, where $w^1 \in \mathbb{R}^6$ is a vector of random perturbations computed at the start of the training process for the initial human policy, with its individual components $w^1_n \sim \mathcal{N}(0, 1)$ for $ n = 1, \dots, 6$, and uses the learning rate value $a = 0.3$. The remaining thinking steps, including levels $j \in \{2, \dots, k^m\}$ for the human agent ($i=h$), and levels $j \in \{1, \dots, k^m\}$ for the robot agent ($i=r$), have initial weights $W_{i,0}^j = [3 \; 3 \; 3 \; 3 \; 3 \; 3]^\top + w^j$, with $w^j_n \sim \mathcal{U}(0.1, 0.5)$ for $ n = 1, \dots, 6$, and use the learning rate value $a = 0.07$. The basis functions are $\phi_i = [x_1^2 \; x_1x_2 \; x_2^2 \; x_3^2 \; x_3x_4 \; x_4^2]^\top $. For all levels we verify the history stack rank condition in \eqref{eq:hist_rank_cond} every $0.1 \; s$ from the start of the level training process, and collect data until the condition is met and the length of the stack is at least $Z=6$.

The objective of each agent in the game is defined by the cooperative running cost $r_c = M(x) + \sum_{j \in \{h, r\}}u_j^\top  R_{j} u_j$ as in \eqref{eq:coop_reward}. For the human objective $r_{c,h} = r_c$, we establish $M(x) = x^\top Qx$. We define the robot objective to be $r_{c,r} = r_{c} / 2$, and $Q = I_4$, $R_h = 2I_2$, $R_r = I_2$. The initial state of the first training rollout is $x_0 = [-1.0 \; 0.7 \; 1.0 \; 0.8]$. The maximum level of rationality being acquired for each agent is $k^m = 5$, which corresponds to $10$ thinking steps in total for the learning procedure. This limit is chosen based on previous works showing that computing up to 5 levels of rationality can be enough to describe human strategic behavior \cite{tan2025human, wright2010beyond, keurulainen2024role}.

The training procedure executes the thinking steps sequentially. The agent acquiring the corresponding rationality level updates its weights throughout the interaction, in which both agents act on the system to achieve the cooperative task. The thinking step runs for a predefined amount of time, enough for the weights to converge to their final value and achieving the task. After completing each thinking step, the final weights values are stored in the bank of policies.

\begin{remark}
    We complement the training procedure by repeating multiple rollouts during each thinking step. Specifically, the duration for the training procedure of a single thinking step is $500$ seconds (simulated time), during which we perform $20$ rollouts. At each rollout $d$ we restart the system's evolution at a randomly picked initial state $x_{0}^{(d)}$ where $x_{0, n}^{(d)} \sim \mathcal{U}(-0.9, 0.9)$, for $n = 1, \dots, 4$.
\end{remark}

\begin{figure}[htb]
    \centering
    \includegraphics[width=\linewidth]{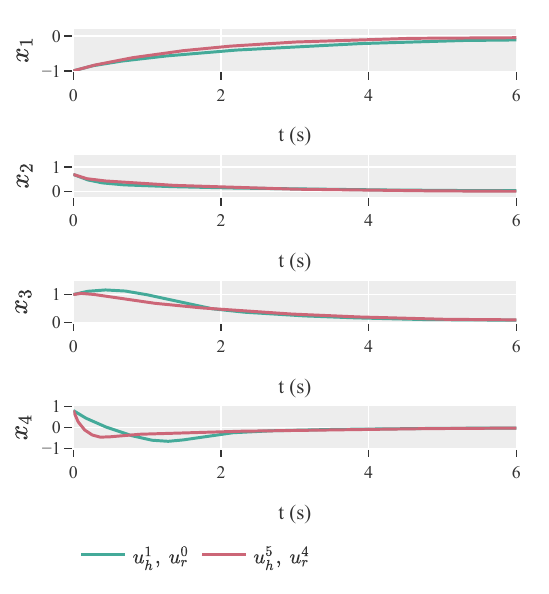}
    \caption{Simulation of the evolution of the state $x$ through time for specific human levels of rationality (using the final training weights). The system's inputs being used are a human level-1 $u_h^1$, and level-5 $u_h^5$, and the robot's response are the corresponding lower level $u_r^{0}$, and $u_r^{4}$ (which are the assumptions from human's perspective while acquiring the level-$k$ policy).}
    \label{fig:level-k_human_state}
\end{figure}

\begin{figure}[htb]
    \centerline{\includegraphics{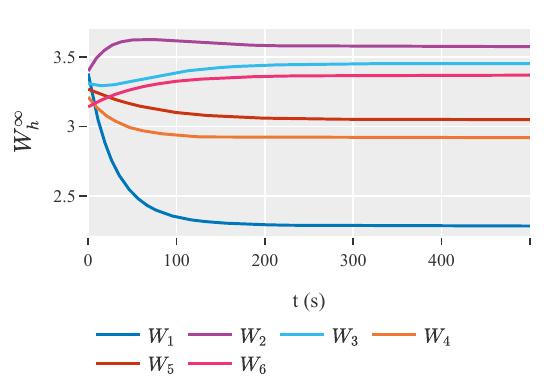}}
    \caption{Weights values over time while acquiring the $u_h^\infty$ rationality level.}
    \label{fig:h_inf_w_converge}
\end{figure}

\begin{figure}[htb]
    \centerline{\includegraphics{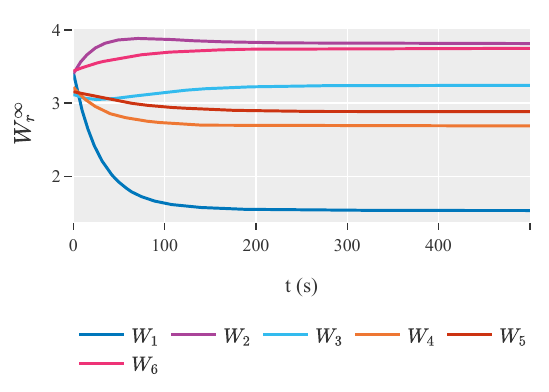}}
    \caption{Weights values over time while acquiring the $u_r^\infty$ rationality level.}
    \label{fig:r_inf_w_converge}
\end{figure}

Figure \ref{fig:level-k_full} shows the final weight values after completing all the thinking steps in the training procedure. These learned weights parameterize the bank of policies used to perform the cooperative task in this setup.

In Figure \ref{fig:level-k_human_state} we show the behavior of the system's state when the agents act on it using different but complementary rationality levels. Both combinations of the agents' bounded rational behaviors accomplish the stabilization task, but exhibit different state trajectories.

The critic-weight trajectories during training are shown in Figure \ref{fig:h_inf_w_converge}, and Figure \ref{fig:r_inf_w_converge} for the thinking step that computes the agents' infinite-level reference policies $u_h^\infty, \; u_r^\infty$.

\begin{figure*}[t!]
    \centerline{\includegraphics{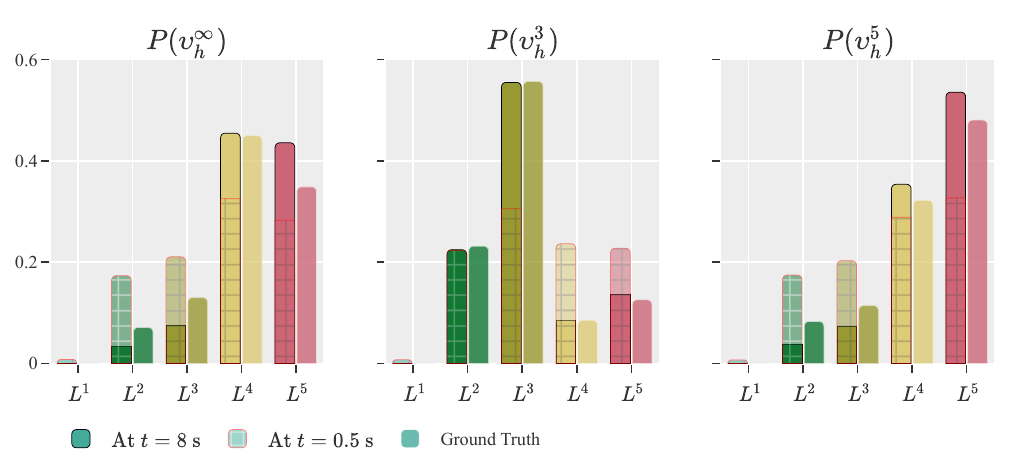}}
    \caption{PHM estimation for three simulated human behaviors, showing each ground-truth distribution and its online estimates at $t$=0.5 s and $t$=8 s.}
    \label{fig:estimation_phm_comparison_eval}
\end{figure*}

\subsubsection{Performing the Shared Control Interaction with an unknown human behavior} \label{sim:perform_sc}

Once the training process ends and we obtain the bank of policies for both agents, we proceed to evaluate our shared control mechanism. For this, we consider the interaction with a \textit{human with unknown behavior}, represented by the policy $u^?_h$. Here, the cooperative goal is achieving the stabilization of (\ref{system_experiment}) having a robot that assists the human control input acting on the system. We do not provide the robot with the simulated human's generating distribution, but assume that the task objective is known and fixed. During the shared control interaction, we follow Algorithm \ref{alg:sc_inter}. We estimate a PHM over the stored level-$k$ candidate policies, as explained in Section \ref{sec:estimating_phm}, and compute the robot assistance using Section \ref{sec:robot_br}.

For comparison and validation purposes, we simulate different unknown human behaviors acting on the system by building their human models as proposed in \cite{tan2025human}. These simulated humans are also based on the level-k policies obtained in the training process and serve as the ground truth PHMs to evaluate our online estimation process.

\begin{remark}
\label{rem:appendix_sol}
Given the experiment's NNs structure, and quadratic basis functions $\phi$ together with forward-Euler state propagation, the one-step optimization admits a closed-form solution. For this experiment, we compute the robot's best response using the closed-form solution derived in Section \ref{sec:br_response_cfsol}. This also allows us to reduce the computational burden of finding the solution at each estimation update of the human model. When computing the robot control action using the best response closed-form solution we verify that the strict-convexity condition $T_{\mathrm{int}}R_r+G_r^\top P G_r\succ0$ is true for the evaluated state and weights.
\end{remark}

We present two evaluation points for the same experiment setup that highlight the relevance of our proposed methods.
\begin{enumerate}
    \item We evaluate the effectiveness of the PHM estimation process by using simulated humans at different levels of rationality.
    \item We compare the proposed robot response with two alternative response rules that also use the estimated PHM.
\end{enumerate}

The duration of the shared control interaction is set to $T=30$ seconds, with $T_{int}=0.06$ seconds, and the integrator step time being $0.01$ seconds. For the estimation of the probabilistic human model, the forgetting factor is set to $\lambda = 1.0$, the inverse-scale parameter to $\beta = 1.2$, and the initial PHM distribution to $\mathcal{P}(\upsilon_h = \upsilon_h^j) = 0.2$ for $j = 1, \dots, k^m$. For the robot best response computation, we use the robot's learned infinite-level weights $W_r^\infty$ to build the cost-to-go value approximation $V_f$. We start applying the robot action when at least $T_{int}$ has passed. The initial state is $x_0 = [3.5 \; 6.7 \; -5.9 \; 4.8]$. We assume that the full state is available for feedback and is measured without error.

\paragraph{Evaluating PHM estimation process} \label{sim:eval_phm_est}

Our PHM estimation process presented in Section \ref{sec:estimating_phm} aims to describe the unknown human behavior in terms of the level-k policies obtained during the learning procedure. To evaluate the effectiveness of our method, we use ground truth distributions $\mathcal{P}_{\text{sim}}(\upsilon_h)$ of level-k human behaviors, and compare our estimation $\mathcal{P}_{\text{PHM}}(\upsilon_h)$ against them using the KL divergence metric over the two distributions as
\begin{equation} \label{eq:exp_kl_div}
    D_{\text{KL}}(\mathcal{P}_{\text{sim}} \space || \space \mathcal{P}_{\text{PHM}})
\end{equation}

To obtain the human behavior distributions we follow the \textit{optimism error} technique detailed in \cite{tan2025human} using the weights in our bank of policies, and compute them before running the evaluations on the shared control interaction. We simulate the system's \eqref{system_experiment} evolution over $T = 30 \; s$, using an integrator time step of $0.01 s$. We consider that the robot's infinite-level reference policy $u_r^\infty$ acts on it, together with a human reference behavior $\upsilon^{k}_h$. This human reference behavior corresponds to a human acting on the system with the candidate policy $u_h^k$, where $k$ is a rationality level previously selected. We repeat this simulation for each human level $j$, for $j = 1, \dots, k^m$, and compute and keep track of $u_h^j$. We establish intervals of interaction of value $T_{int} = 0.06$ during this period. The optimism error $o$ for a specific level $j$ at the interval $\ell$ is $o_\ell^j = \int_{T_{\text{int}}} \lVert u_h^k - u_h^j \rVert_2 d\tau$. The vector $\bm{O} = [o^1, \dots, o^{k^m}]$ contains the accumulated optimism errors for all the intervals and all the levels. We can obtain the PHM describing a reference human behavior $\mathcal{P}(\upsilon_h = \upsilon^{k}_h)$ with a predominant level $k$ of rationality as
\begin{equation} \label{eq:optim_sim_models}
    \mathcal P\!\left(\upsilon_h=\upsilon_h^k\mid\bm O\right)
    =\frac{\exp\!\left(-\beta o^j\right)}
    {\sum_{q=1}^{k^m}\exp\!\left(-\beta o^q\right)}.
\end{equation}

Concretely, we use \eqref{eq:optim_sim_models} to define the following distributions and use them to simulate human behavior during our shared control interaction:
\begin{enumerate}
    \item An \textbf{optimistic} human model: $\mathcal{P}_{\text{sim}}(\upsilon_h = \upsilon_h^{\infty})$. This is the human model obtained with the \textit{optimism error} technique, where the comparison is against the $u_h^{\infty}$. This model remains a distribution over the finite level-1--level-5 candidate bank; $u_h^\infty$ is the optimism-error reference.
    \item A predominantly \textbf{level-3} human model: $\mathcal{P}_{\text{sim}}(\upsilon_h =\upsilon_h^{3})$. Here, the comparison in the \textit{optimism error} is against the $u_h^{3}$.
    \item A predominantly \textbf{level-5} human model: $\mathcal{P}_{\text{sim}}(\upsilon_h =\upsilon_h^{5})$. Here, the optimism-error comparison is against $u_h^{5}$.
\end{enumerate}

Additionally, Figure \ref{fig:estimation_phm_comparison_eval} presents the results of the PHM estimation process at two reported times during the interaction, one close to the beginning of it (at $t = 0.5$ s), and another where more time has passed and more measurements have been taken (at $t = 8$ s). These results show how the estimated PHM evolves through time, and gets closer to the ground truth distribution governing the human input behavior. Table \ref{tab:exp_phm_gt_dist} shows the levels' probabilities in each reference PHM distribution $\mathcal{P}_{\text{sim}}(\upsilon_h)$ used for evaluating the estimation process. The predominant level $k$ is the one with the highest probability in its corresponding distribution $\mathcal{P}_{\text{sim}}(\upsilon_h = \upsilon^k_ h)$. Below each reference distribution, Table \ref{tab:exp_phm_gt_dist} also shows the probability mass for each level-k in all the estimated PHM distributions $\mathcal{P}_{\text{PHM}}(\upsilon_h)$ at $t = 8$ s. When estimating $\mathcal{P}_{\text{sim}}(\upsilon_h^{3})$, the estimated PHM $\mathcal{P}_{\text{PHM}}(\upsilon_h^{3})$ assigns its largest probability to the predominant level with probability mass closer to the ground-truth $L^3$, which results in a lower KL divergence. Similarly, the estimated PHM $\mathcal{P}_{\text{PHM}}(\upsilon_h^{\infty})$ assigns the predominant level probability mass closer to the ground-truth $L^4$, but a much larger probability mass to $L^5$ than the groun-truth value, increasing the KL divergence measurement. Finally, the estimated PHM $\mathcal{P}_{\text{PHM}}(\upsilon_h^{5})$ also assigns a larger probability mass to its predominant level than the ground-truth probability. The KL divergence values in Table \ref{tab:exp_phm_kl_div} decrease at the reported times for all three scenarios. The level-$\infty$ and level-5 scenario retain the higher endpoints divergence ($0.042$ and $0.036$, compared with $0.002$), indicating that they are the most difficult of the three cases in this benchmark.

\begin{table}[h!tb]
\begin{tabular*}{\columnwidth}{@{\extracolsep\fill}clllll@{}}
\toprule
     & \multicolumn{5}{c}{$L^k$ Probabilities} \\
    PHM Distribution & $L^1$ & $L^2$ & $L^3$ & $L^4$ & $L^5$ \\
\midrule
    $\mathcal{P}_{\text{sim}}(\upsilon_h =\upsilon_h^\infty)$ & $2\times10^{-4}$ & $0.007$ & $0.130$ & $\mathbf{0.449}$ & $0.349$ \\
    $\mathcal{P}_{\text{PHM}}(\upsilon_h =\upsilon_h^\infty)$ & $4\times10^{-7}$ & $0.003$ & $0.008$ & $\mathbf{0.455}$ & $0.436$ \\
\midrule
    $\mathcal{P}_{\text{sim}}(\upsilon_h = \upsilon_h^3)$ & $3\times10^{-4}$ & $0.232$ & $\mathbf{0.557}$ & $0.008$ & $0.126$ \\
    $\mathcal{P}_{\text{PHM}}(\upsilon_h = \upsilon_h^3)$ & $1\times10^{-6}$ & $0.225$ & $\mathbf{0.555}$ & $0.008$ & $0.136$ \\
 \midrule
    $\mathcal{P}_{\text{sim}}(\upsilon_h=\upsilon_h^5)$ & $2\times10^{-4}$ & $0.008$ & $0.114$ & $0.322$ & $\mathbf{0.481}$ \\
    $\mathcal{P}_{\text{PHM}}(\upsilon_h=\upsilon_h^5)$ & $4\times10^{-7}$ & $0.004$ & $0.007$ & $0.354$ & $\mathbf{0.536}$ \\
\bottomrule
\end{tabular*}
\caption{Probabilities of each level $L^k$ for all the ground truth distributions $\mathcal{P}_{\text{sim}}$ used to evaluate the PHM estimation process, and all the estimated distributions $\mathcal{P}_{\text{PHM}}$ at $t = 8 \;s$.}
\label{tab:exp_phm_gt_dist}
\end{table}

\begin{table}[h!tb]
\begin{tabular*}{\columnwidth}{@{\extracolsep\fill}clll@{}}
\toprule
    & \multicolumn{3}{c}{KL divergence} \\
    Simulated Human Input & $t = 0$ s & $t = 0.5$ s & $t = 8$ s \\
\midrule
    $\mathcal{P}_{\text{sim}}(\upsilon_h^\infty)$ & $0.428$ & $0.092$ & $\mathbf{0.042}$ \\
\midrule
    $\mathcal{P}_{\text{sim}}(\upsilon_h^3)$ & $0.471$ & $0.180$ & $\mathbf{0.002}$ \\
 \midrule
    $\mathcal{P}_{\text{sim}}(\upsilon_h^5)$ & $0.437$ & $0.093$ & $\mathbf{0.036}$ \\
\bottomrule
\end{tabular*}
\caption{KL divergence between the ground truth models and the estimated PHM at different times during the shared control interactions.}
\label{tab:exp_phm_kl_div}
\end{table}

\paragraph{Evaluating the Robot's best response} \label{sim:eval_rbr}

The second evaluation aims to compare our computation proposal for the robot response $u_r^{\text{br,}\star}$ against other two alternatives. We run the same shared control interaction for 3 different robot responses. One is our proposed method from Section \ref{sec:robot_br}. While, the other two alternatives are:
\begin{enumerate}
    \item $u_r^{\text{br, max}}$ - The robot's level-k response to the most likely human behavior in the estimated PHM.
    \item $u_r^{\text{br, avg}}$ - A weighted average of all the robot's level-k responses in the bank of policies. The weights for the average correspond to the probabilities of each level in the estimated PHM.
\end{enumerate}

At the start of the $T_{\text{int}}$ interval, the simulated human model samples a level-$k$ policy from the ground truth PHM distribution and uses the bank of policies to compute the control input according to that level-$k$ policy during the $T_{\text{int}}$. At the same time the concrete robot response being evaluated is obtained using the most recent update of the estimated PHM by solving the minimization problem described in \eqref{eq:robot_best_resp}, or one of the alternative responses. We choose to measure the control effort for each agent $i$ over the horizon $T$ as $J_{u_i}=\int_0^T u_i^\top R_i u_i\,d\tau$, and the finite-horizon cost for the robot $J_{r,T}=\int_0^T r_{c,r}\,d\tau$ to further analyze the agents' contributions and compare the alternatives baselines. For all the robot response alternatives we use the same initial state, and simulate human behavior using the optimistic human model distribution $\mathcal{P}_{\text{sim}}(\upsilon_h = \upsilon_h^\infty)$.

\begin{table}[htpb]
\begin{tabular*}{\columnwidth}{@{\extracolsep\fill}llll@{}}
\toprule
    & \multicolumn{3}{c}{Robot Response Alternatives} \\
    Control Effort & \bm{$u_r^{\text{br,}\star}$} & $u_r^{\text{br, max}}$ & $u_r^{\text{br, avg}}$ \\
\midrule
    $J_{u_r}$ & $25.79$ & $182.03$ & $122.04$ \\
    $J_{u_h}$ & $67.56$ & $64.91$ & $62.47$ \\
\bottomrule
\end{tabular*}
\caption{Accumulated Control Effort over the shared control interaction for each robot response alternative.}
\label{tab:total_effort_rbr_alt}
\end{table}

\begin{table}[h!tpb]
\begin{tabular*}{\columnwidth}{@{\extracolsep\fill}llll@{}}
\toprule
    & \multicolumn{3}{c}{Robot Response Alternatives} \\
    & \bm{$u_r^{\text{br,}\star}$} & $u_r^{\text{br, max}}$ & $u_r^{\text{br, avg}}$ \\
\midrule
    $J_{r,T}$ & $\bm{112.42}$ & $187.62$ & $155.05$ \\
\bottomrule
\end{tabular*}
\caption{Finite-horizon cost $J_{r,T}$ over the period $T$ for the robot agent when performing the shared control interaction using each robot response alternative.}
\label{tab:total_cost_rbr_alt}
\end{table}

\begin{figure}[htb]
    \centering
    \includegraphics[width=\linewidth]{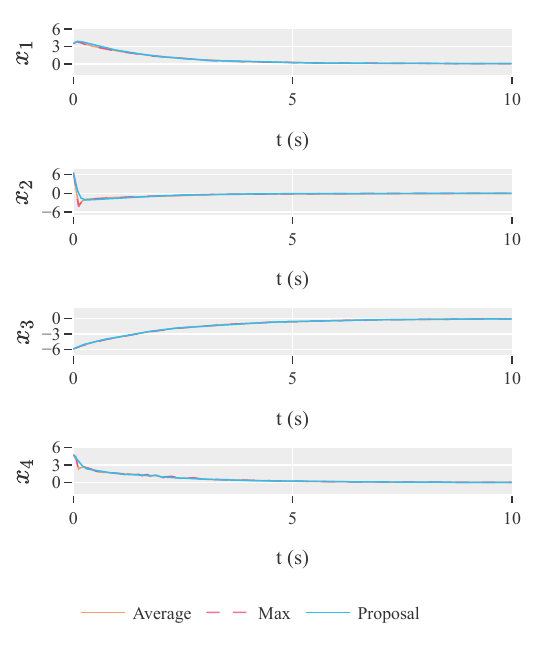}
    \caption{System's state evolution when using each of the different robot response alternatives during the shared control interaction.}
    \label{fig:robot_br_states}
\end{figure}

\begin{figure}[htb]
    \centering
    \includegraphics[width=\linewidth]{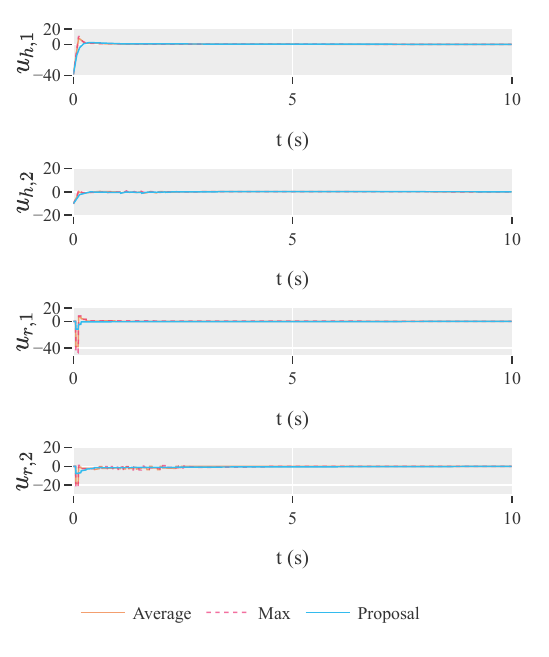}
    \caption{Human ($u_h$) and robot ($u_r$) control input trajectories under the three robot-response rules during the shared control interaction.}
    \label{fig:robot_br_inputs}
\end{figure}

\begin{table}[h!tpb]
\begin{tabular*}{\columnwidth}{@{\extracolsep\fill}lll@{}}
\toprule
    & \multicolumn{2}{c}{Statistics (50 samples)} \\
    Measurements & Mean & Standard Deviation \\
\midrule
    $D_{\text{KL}}(\mathcal{P}_{\text{sim}}^{\infty} \space || \space \mathcal{P}_{\text{PHM}})$ at $t = 8$s & $0.057$ & $0.051$ \\
    $J_{r,T}$ & $91.736$ & $53.463$ \\
\bottomrule
\end{tabular*}
\caption{Statistics for a sample of 50 different initial states when performing the shared control interaction process in Section \ref{sec:performing_sc_int} using the robot's best response $u_r^{\text{br}, \star}$. The measurements are the KL divergence $D_{\text{KL}}(\mathcal{P}_{\text{sim}}^{\infty} \space || \space \mathcal{P}_{\text{PHM}})$, and the finite-horizon cost $J_{r,T}$ for the robot agent.}
\label{tab:stats_total_effort_rbr_alt}
\end{table}

\begin{figure}[htb]
    \centering
    \includegraphics[width=\linewidth]{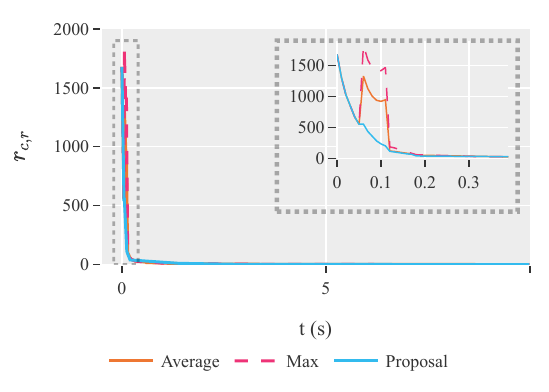}
    \caption{Robot running-cost trajectories under the three robot-response rules during the shared control interaction.}
    \label{fig:robot_br_costs}
\end{figure}

Figures \ref{fig:robot_br_states}, \ref{fig:robot_br_inputs}, \ref{fig:robot_br_costs} show results from the shared control interaction using each of the robot response alternatives. Figure \ref{fig:robot_br_states} shows the state evolution for each of the response alternatives. All of them achieve the stabilization goal but there are clear differences on the trajectories they follow between the ones using the bank of policies directly ($u_r^{\text{br, max}}$, $u_r^{\text{br, avg}}$), and our proposal $u_r^{\text{br,}\star}$. Figure \ref{fig:robot_br_inputs} contains the control input values over the time of the interaction. The two top graphs show the simulated human control action following the ground truth distribution. While, the two at the bottom are the robot responses coming from the different alternative computations. Our proposed method shows lower values overall when compared to the other two alternatives. Together, the human- and robot-input plots show how the response rules redistribute control effort between the agents. Table \ref{tab:total_effort_rbr_alt} provides the measurements of control effort for each agent, showing that the proposed response uses the least robot effort but the greatest human effort.
Relative to the maximum-probability and probability-weighted baselines, the proposed response reduces the accumulated running cost by $40.1\%$ and $27.5\%$, respectively. Its total human-plus-robot control effort is also lower, although this result reflects a marked redistribution toward human effort.
In addition, Figure \ref{fig:robot_br_costs} shows the value of $r_c$ during each step in the interaction. The values coming from the runs using the alternatives based on the bank of polices ($u_r^{\text{br, max}}$, $u_r^{\text{br, avg}}$) have two additional peaks influenced by the increased effort exhibit by the robot input at those times, whereas our proposal shows a smoother behavior. Table \ref{tab:total_cost_rbr_alt} complements this result by comparing the total cost accumulated over the interaction, presenting the lowest value when using our proposal as the robot response computation method. Ultimately, Table \ref{tab:stats_total_effort_rbr_alt} contains summary statistics of the KL divergence metric between the estimated PHM $\mathcal{P}_{\text{PHM}}$ and the simulated human using the optimistic PHM $\mathcal{P}_{\text{sim}}^{\infty}$ using \eqref{eq:exp_kl_div}, and the robot's finite-horizon cost $J_{r,T}$ measurement when performing the shared control interaction procedure for 50 different initial states with the robot agent using the distribution-aware one-step optimization solution $u_r^{\text{br}, \star}$.
Across these runs, the endpoint KL divergence is $0.057\pm0.051$ and the accumulated running cost is $91.736\pm53.463$ (mean $\pm$ standard deviation).

\subsection{Planar Manipulator System}

\subsubsection{System Description}

\begin{figure}[h!tb]
    \centering
    \includegraphics[width=\linewidth]{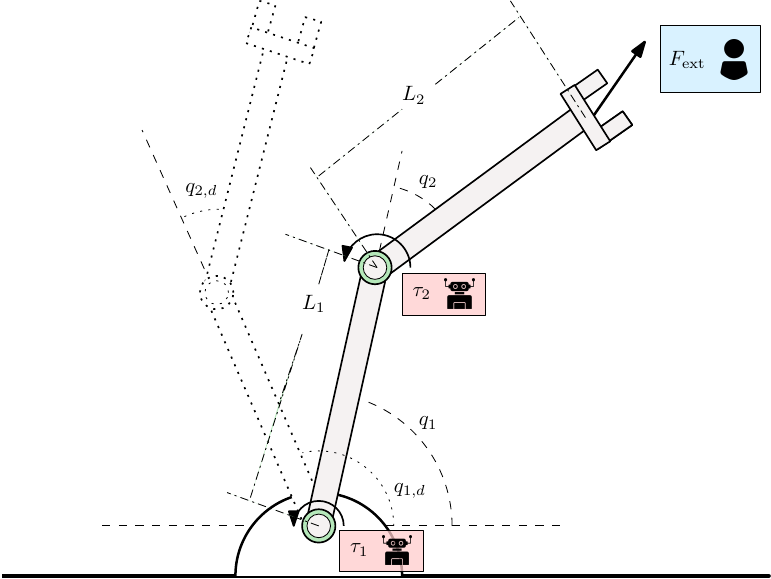}
    \caption{Diagram of the planar manipulator system under a shared control interaction setup. The dotted manipulator represents the placement with the desired joints configuration. The joints torques are controlled by the robot agent, while the external force applied to the end-effector is controlled by the human agent.}
    \label{fig:pr_arm_diagram}
\end{figure}

For this simulation experiment, we use a 2-link robotic manipulator system with revolute joints operating on a two dimensional plane. Two players act on the system to achieve the same cooperative goal. A human exerts a force to the manipulator's end-effector, and a robot actuates the joints' torques. By applying their corresponding control, the players seek to place the joints' angles in a desired configuration. Figure \ref{fig:pr_arm_diagram} shows a diagram of the system's components, and indicates the points of impact for the human and robot agent.

\begin{figure*}[t]
    \centerline{\includegraphics{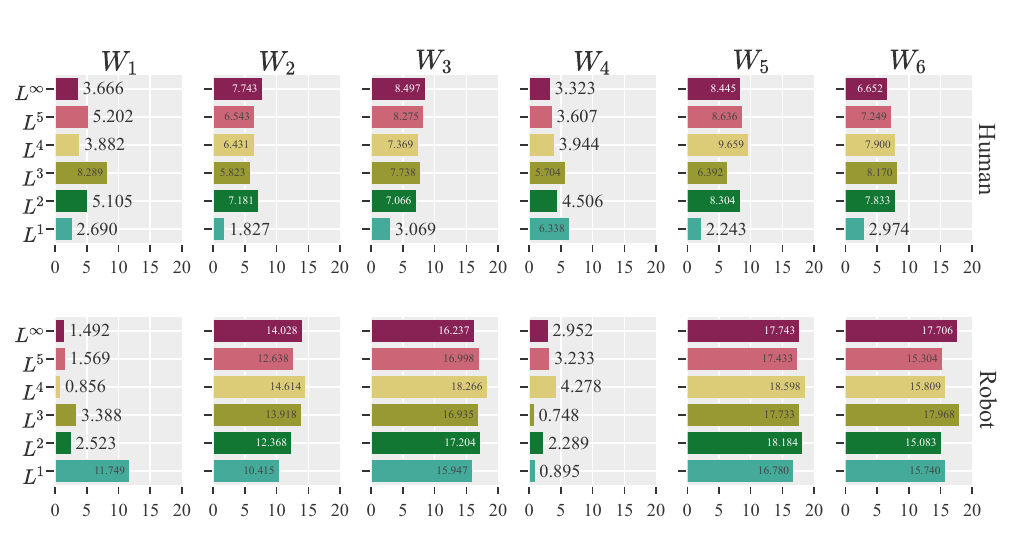}}
    \caption{Bank of Policies up to level-5 for the cooperative control of the planar robotic manipulator. Top row shows the final weights for each level of human rationality, similarly the bottom row shows those for the robot agent. The weights of the NN neurons correspond to each column.}
    \label{fig:pr_arm_level-k_full}
\end{figure*}

The dynamics of the planar manipulator are given by

\begin{equation}
    H(q) \ddot{q} + C(q, \dot{q}) \dot{q} = \tau + J(q)^T F_{\text{ext}}
\end{equation}

where $q \in \mathbb{R}^{2}$ is the joint-angle vector, and $\dot{q}$ and $\ddot{q}$ are the joints-velocity and joints-acceleration vectors, respectively. $H(\cdot) \in \mathbb{R}^{2\times2}$ is the system's inertia matrix, $C(\cdot) \in \mathbb{R}^{2\times2}$ is the Coriolis and centrifugal force matrix, and $J(\cdot) \in \mathbb{R}^{2\times2}$ is the system's Jacobian matrix. $\tau \in \mathbb{R}^{2}$ represents the torques applied to the joints' actuators, and $F_{\text{ext}} \in \mathbb{R}^{2}$ is an external force applied to the system's end-effector. The effects of gravity on the system's dynamics are neglected. We define the $H(\cdot)$, and $C(\cdot)$ matrices to be
\begin{align*}
    H_{1, 1} &= (m_1 + m_2) L_1^2 + m_2 L_2^2 + 2 m_2 L_1 L_2 \cos{q_2} \\
    H_{1, 2} &= H_{2, 1} = m_2 L_2^2 + m_2 L_1 L_2 \cos{q_2} \\
    H_{2, 2} &= m_2 L_2^2 \;,
\end{align*}
\begin{align*}
    c &= m_2 L_1 L_2 \sin{q_2} \\
    C_{1, 1} &= -c \: \dot{q}_2 \\
    C_{1, 2} &= - c \: (\dot{q_1} + \dot{q_2}) \\
    C_{2, 1} &= c \: \dot{q_1} \\
    C_{2, 2} & = 0 \; ,
\end{align*}

and the system's Jacobian matrix $J(\cdot)$ as
\begin{align*}
    J_{1, 1} &= -L_1 \sin(q_1) -L_2 \sin(q_1+q_2) \\
    J_{1, 2} &= -L_2 \sin(q_1+q_2) \\
    J_{2, 1} &= L_1 \cos(q_1) + L_2 \cos(q_1+q_2) \\
    J_{2, 2} & = L_2 \cos(q_1+q_2) \; ,
\end{align*}

where $m_1 = m_2 = 1$ kg are the masses at the joint centers (with the link masses being negligible), and $L_1 = 0.5$, and $L_2 = 0.6$ are the lengths of link 1, and 2 respectively. For all simulations we use an integrator step size of $0.01 \; s$.

We define the system's state variable $x \in \mathbb{R}^{4}$ to be
\begin{equation}
    x = \begin{bmatrix}
        e_{q_1} \\ \dot{q}_1 \\ e_{q_2} \\ \dot{q}_2
    \end{bmatrix}
\end{equation}
where $e_{q_n} = q_n - q_{n, d}, n \in \{1 ,2 \}$ is the joint's error value, and $q_{n, d}$ is the desired configuration for the n-th joint. Following our Human-Robot cooperative system formulation in Section \ref{sec:hr-coop-sys}, we use the state-space control-affine model representation for our system as
\begin{equation}
    \dot{x} = f(x) + g_{F_{\text{ext}}}(x) F_{\text{ext}} + g_{\tau}(x) \tau
\end{equation}

where $f(x)$ is the drift dynamics of the system, and $g_{F_{\text{ext}}}$, and $g_{\tau}$ are the input matrices for their corresponding agents' contribution. $F_{\text{ext}}$ corresponds to the human control input $u_h$, and ${\tau}$ to the robot control input $u_r$.

In this shared control interaction the human agent applies the \textit{external force} to the end-effector of the robotic manipulator aiming to place the manipulator joints in a desired configuration. Simultaneously, the robot control input acts as an assistive response modifying the joints' torques in order to achieve the same cooperative task. For the experiments described in this section we establish the desired joints' configuration to be $q_d = [ -0.5 \; -0.7 ]$. The task's goal is to converge the state $x$ to $\mathbf{0}$, which will indicate that the current joints' configuration $q$ have reached the desired configuration $q_d$ and remain there with $\dot{q} = \mathbf{0}$. The selected initial joint configuration is $q_1 = 0.9$, and $q_2=0.8$, resulting in the initial state $x_0 = [1.4 \; 0 \; 1.5 \; 0]$.

\subsubsection{Learning the rationality levels model}

First, using Algorithm \ref{alg:level_k_model} we obtain the level-k candidate policies representing the players' potential behaviors, up to the maximum level $k^m = 5$. For each player $i \in \{h ,r\}$, we acquire the levels $j$, for $j \in \{1, \ldots, k^m \}$. The neural network approximating the value functions consists of $6$ neurons, with the corresponding quadratic basis functions being $\phi_i = [x_1^2 \; x_1x_2 \; x_2^2 \; x_3^2 \; x_3x_4 \; x_4^2]^\top $. When acquiring the initial human policy in \ref{sec:init_h_policy} we use the initial weights $W_{h,0}^1 = [3 \; 3 \; 3 \; 3 \; 3 \; 3] + w^1$ with $w^1_n \sim \mathcal{N}(0, 1)$, for $n \in \{1, \ldots, 6\}$. For the remaining human levels $j$, for $j \in \{2, \ldots, k^m\}$, we use $W_{h,0}^j = [8 \; 8 \; 8 \; 8 \; 8 \; 8] + w^j$ with $w^j_n \sim \mathcal{N}(0, 1)$, for $n \in \{1, \ldots, 6\}$. For the robot levels $j$, for $j \in \{1, \ldots, k^m\}$, the initial weights are $W_{r,0}^j = [16 \; 16 \; 16 \; 16 \; 16 \; 16] + w^j$ with $w^j_n \sim \mathcal{N}(0, 1)$, for $n \in \{1, \ldots, 6\}$. The perturbation $w^j$ added to each set of initial weights is computed at the start of the corresponding thinking step. The learning rate for all steps is $a = 0.015$. We start recording samples for the history stack after $2$ seconds, every $0.01$ seconds, and stop until condition \eqref{eq:hist_rank_cond} is met and the length of the history stack is at least $Z = 6$. We use the running cost $r_c = M(x) + \sum_{i \in \{h, r\}}u_i^\top  R_{i} u_i$ as in \eqref{eq:coop_reward} to define each agent's objective. For the human agent $r_{c, h} = r_c$, with $M(x) = x^\top Qx$, and for the robot $r_{c,r} = r_{c, h} / 2$, where $Q = I_4$, $R_h = 30I_2$, $R_r = 150I_2$. The total training time for each thinking step is $T = 600$.

\begin{figure}[htb]
    \centering
    \includegraphics[width=\linewidth]{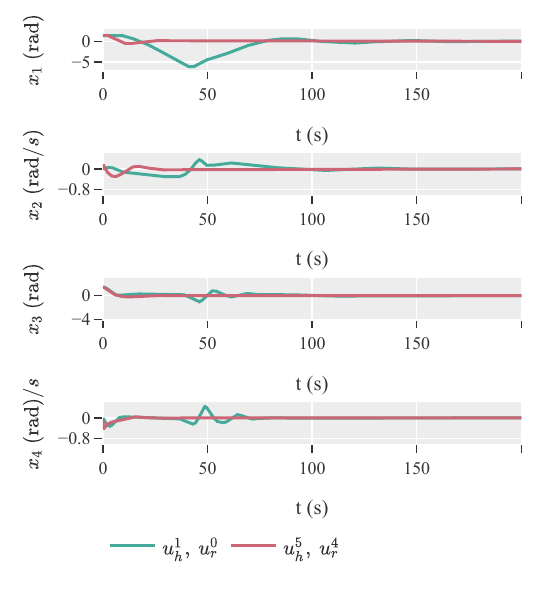}
    \caption{Evolution of the state $x$ through time while learning the human level of rationality 1 and 5. While learning the human level-1 $u_h^1$, the fixed robot response is $u_r^{0}$ (which corresponds to no robot contribution), and while learning the level-5 $u_h^5$, the corresponding fixed lower robot level is $u_r^{4}$.}
    \label{fig:pr_arm_level-k_human_state}
\end{figure}

\begin{figure}[htb]
    \centerline{\includegraphics{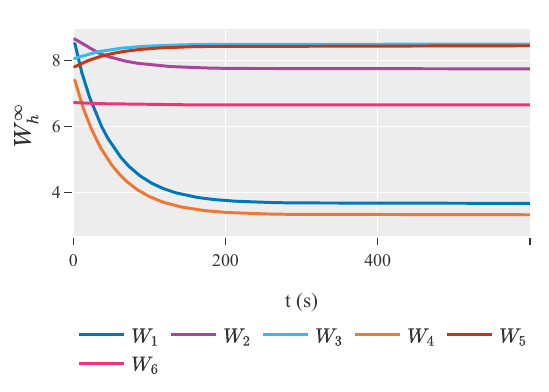}}
    \caption{Weights evolution while acquiring the $u_h^\infty$ rationality level during the planar robotic manipulator experiment.}
    \label{fig:pr_arm_h_inf_w_converge}
\end{figure}

\begin{figure}[htb]
    \centerline{\includegraphics{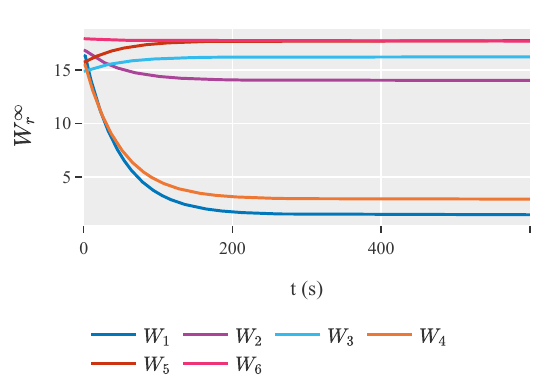}}
    \caption{Weights evolution while acquiring the $u_r^\infty$ rationality level during the planar robotic manipulator experiment.}
    \label{fig:pr_arm_r_inf_w_converge}
\end{figure}

The resulting bank of policies at the end of the iterative training procedure is represented in Figure \ref{fig:pr_arm_level-k_full}. Each plot in Figure \ref{fig:pr_arm_level-k_human_state} shows how the state variables evolve during the acquisition of the initial human policy $u_h^1$ when no robot contribution is being assumed by the human agent, and the human policy of level-5 $u_h^5$ that assumes the robot contribution is of one level lower $u_r^4$. Both policy pairs complete the stabilization task but produce different transient trajectories. Figure \ref{fig:pr_arm_h_inf_w_converge} and Figure \ref{fig:pr_arm_r_inf_w_converge} show the critic-weight trajectories for the infinite-level reference policies ($u_h^\infty$, $u_r^\infty$).

\subsubsection{Performing the Shared Control Interaction with an unknown human behavior} \label{sim:pr_arm_perform_sc}

Following the acquisition of the level-k bank of policies, we perform the shared control of the planar robotic manipulator considering the control actions of a human with unknown rational behavior $u_h^?$, and a robot responding based on its estimation of the human behavior following the methods presented in Section \ref{sec:performing_sc_int}.

Similarly to the previous experiment in Section \ref{sim:perform_sc}, we present the results of applying our proposed methods in this practical setup by separating the analysis on two parts. First, the effectiveness of the PHM estimation to capture the ground-truth PHM. Second, the performance of the robot's response when using the one-step distribution-aware solution compared to the other alternative baselines using the estimated PHM.

The total time for the shared control interaction is $T = 180$ seconds, with the duration of the interval of interaction $T_{\text{int}} = 0.1$ s. The initial PHM distribution is set to $\mathcal{P}(\upsilon_h = \upsilon_h^j) = 0.2$, for $j \in \{1, \ldots, k^m \}$, the forgetting factor is $\lambda = 1.0$, the inverse-scale parameter is $\beta = 1$ and we use the weights of the infinite-reference robot policy $W_r^\infty$ to approximate the cost-to-go value function $V_f$.

\paragraph{Evaluating PHM estimation process}

We follow the same procedure as in the previous experiment section \ref{sim:eval_phm_est} and simulate the unknown human behavior $u_h^?$ with three ground-truth PHM distributions ($\mathcal{P}_{\text{sim}}(\upsilon_h^{\infty})$, $\mathcal{P}_{\text{sim}}(\upsilon_h^{3})$, and $\mathcal{P}_{\text{sim}}(\upsilon_h^{5})$) obtained using the optimism error technique to compute \eqref{eq:optim_sim_models}. For each we compute our PHM estimation $\mathcal{P}_{\text{PHM}}(\upsilon_h)$, and measure their comparison using the KL divergence metric as in \eqref{eq:exp_kl_div}.

\begin{figure*}[t!]
    \centerline{\includegraphics{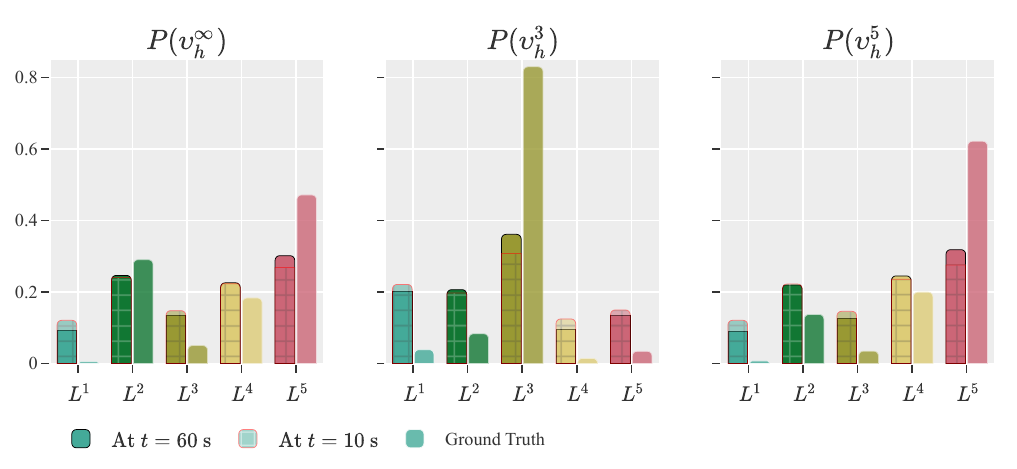}}
    \caption{PHM estimation process evaluation for the shared control of the planar manipulator system. For the three simulated human behaviors we show the ground truth PHM, and the corresponding estimation at two different times during the process.}
    \label{fig:pr_arm_estimation_phm_comparison_eval}
\end{figure*}

A summary of the comparison is presented in Figure \ref{fig:pr_arm_estimation_phm_comparison_eval}. The estimated PHM $\mathcal{P}_{\text{PHM}}(\upsilon_h)$ is presented first at a time closer to the start of the interaction ($t = 10$ seconds), and then its evolution once the interaction has progressed ($t = 60$ seconds). For the three simulated scenarios, the predominant level in the estimation matches the ground truth, although their probability mass is lower than the corresponding ground-truth value. Table \ref{tab:pr_arm_exp_phm_gt_dist} shows the probability values for the ground truth distributions, and the estimation at $t=60 \: s$. Given that the simulated PHMs use the policy space to compute the difference the levels, and we use the state residuals, some difference between the probabilistic models is expected. For all scenarios, the estimated PHM follows a similar shape to the ground-truth ones. Table \ref{tab:pr_arm_exp_phm_kl_div} describes the comparison numerically using the KL divergence metric. The divergence decreases at the reported times in all three cases. However, the level-3 case remains the least accurate at $t=60$ s, with $D_{\mathrm{KL}}=0.476$, compared with $0.156$ and $0.248$ for the infinite-level and level-5 reference scenarios.

\begin{table}[h!tb]
\begin{tabular*}{\columnwidth}{@{\extracolsep\fill}clllll@{}}
\toprule
     & \multicolumn{5}{c}{$L^k$ Probabilities} \\
    PHM Distribution & $L^1$ & $L^2$ & $L^3$ & $L^4$ & $L^5$ \\
\midrule
    $\mathcal{P}_{\text{sim}}(\upsilon_h =\upsilon_h^\infty)$ & $0.005$ & $0.290$ & $0.050$ & $0.183$ & $\mathbf{0.471}$ \\
    $\mathcal{P}_{\text{PHM}}(\upsilon_h =\upsilon_h^\infty)$ & $0.092$ & $0.246$ & $0.134$ & $0.226$ & $\mathbf{0.302}$ \\
\midrule
    $\mathcal{P}_{\text{sim}}(\upsilon_h = \upsilon_h^3)$ & $0.039$ & $0.083$ & $\mathbf{0.829}$ & $0.014$ & $0.034$ \\
    $\mathcal{P}_{\text{PHM}}(\upsilon_h = \upsilon_h^3)$ & $0.201$ & $0.207$ & $\mathbf{0.361}$ & $0.096$ & $0.135$ \\
 \midrule
    $\mathcal{P}_{\text{sim}}(\upsilon_h=\upsilon_h^5)$ & $0.007$ & $0.137$ & $0.035$ & $0.199$ & $\mathbf{0.622}$ \\
    $\mathcal{P}_{\text{PHM}}(\upsilon_h=\upsilon_h^5)$ & $0.090$ & $0.219$ & $0.127$ & $0.245$ & $\mathbf{0.318}$ \\
\bottomrule
\end{tabular*}
\caption{Probabilities of each level $L^k$ for all the ground truth distributions $\mathcal{P}_{\text{sim}}$ used to evaluate the PHM estimation process, and all the estimated distributions $\mathcal{P}_{\text{PHM}}$ at $t = 60 \;s$.}
\label{tab:pr_arm_exp_phm_gt_dist}
\end{table}

\begin{table}[h!tb]
\begin{tabular*}{\columnwidth}{@{\extracolsep\fill}clll@{}}
\toprule
    & \multicolumn{3}{c}{KL divergence} \\
    Simulated Human Input & $t = 0$ s & $t = 10$ s & $t = 60$ s \\
\midrule
    $\mathcal{P}_{\text{sim}}(\upsilon_h^\infty)$ & $0.409$ & $0.215$ & $\mathbf{0.156}$ \\
\midrule
    $\mathcal{P}_{\text{sim}}(\upsilon_h^3)$ & $0.946$ & $0.602$ & $\mathbf{0.476}$ \\
 \midrule
    $\mathcal{P}_{\text{sim}}(\upsilon_h^5)$ & $0.569$ & $0.337$ & $\mathbf{0.248}$ \\
\bottomrule
\end{tabular*}
\caption{KL divergence measurement for each comparison between the ground truth models and the estimated PHM at different times during the shared control interactions with the planar manipulator system.}
\label{tab:pr_arm_exp_phm_kl_div}
\end{table}

\paragraph{Evaluating the Robot's best response}

Similarly to Section \ref{sim:eval_rbr}, we compare $u_r^{\text{br,}\star}$, against the $u_r^{\text{br,}\text{max}}$, and $u_r^{\text{br,}\text{avg}}$ alternatives. At the start of the $T_{\text{int}}$ we sample the rationality level defining the candidate policy to use for the simulated human from the reference distribution, and use the most recent estimation of the PHM to compute the control action for the robot response alternative. We compute the finite-horizon cost for the robot $J_{r, T} = \int_0^T r_{c,r} \: d\tau$ as the metric to compare the alternatives performance.

\begin{figure}[h!tb]
    \centering
    \includegraphics[width=\linewidth]{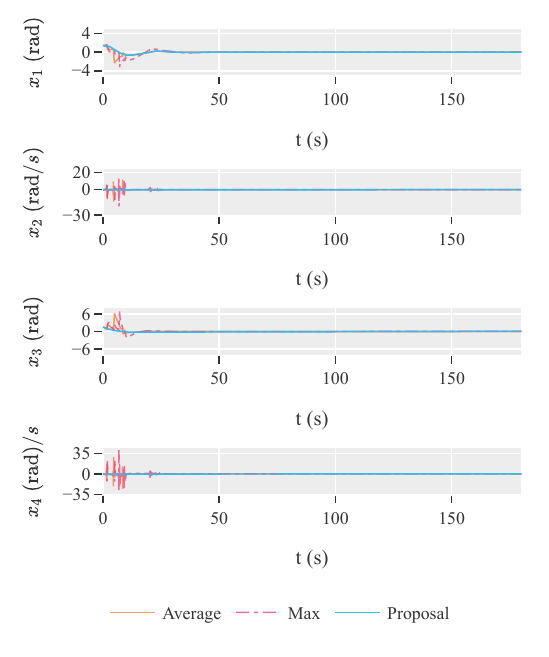}
    \caption{Planar Manipulator System's state evolution when applying each of the different robot response alternatives during the shared control interaction.}
    \label{fig:pr_arm_robot_br_states}
\end{figure}

\begin{figure}[h!tb]
    \centering
    \includegraphics[width=\linewidth]{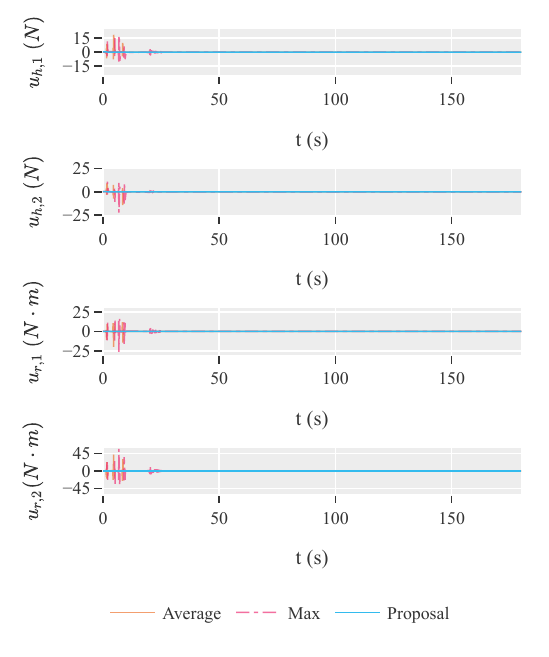}
    \caption{Human (external force $u_h$) and robot (joints' torques $u_r$) control inputs obtained by using each of the different robot response alternatives during the shared control interaction.}
    \label{fig:pr_arm_robot_br_inputs}
\end{figure}

\begin{figure}[h!tb]
    \centering
    \includegraphics[width=\linewidth]{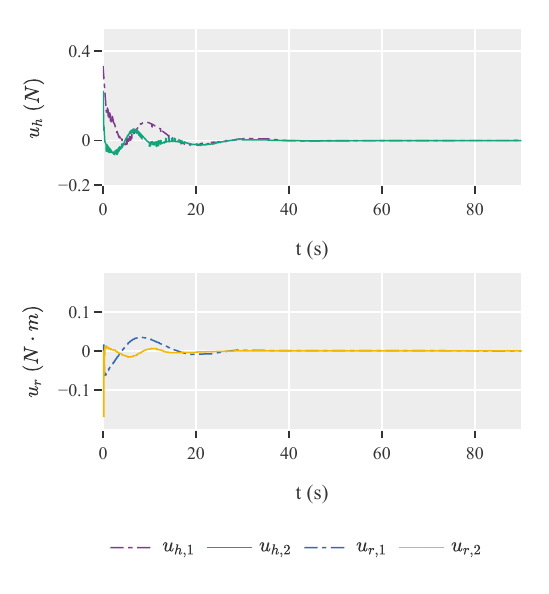}
    \caption{Human $u_h$ and robot $u_r$ control inputs over time for the manipulator shared control interaction, when the robot uses the distribution-aware one-step optimization solution to assist the human operator.}
    \label{fig:pr_arm_robot_rbr_only_inputs}
\end{figure}

\begin{table}[h!tpb]
\begin{tabular*}{\columnwidth}{@{\extracolsep\fill}llll@{}}
\toprule
    & \multicolumn{3}{c}{Robot Response Alternatives} \\
    & \bm{$u_r^{\text{br,}\star}$} & $u_r^{\text{br, max}}$ & $u_r^{\text{br, avg}}$ \\
\midrule
    $J_{r, T}$ & $\bm{11.10}$ & $977.5 \times10^{3}$ & $221.11 \times10^{3}$ \\
\bottomrule
\end{tabular*}
\caption{Finite-horizon cost $J_{r, T}$ of the robot agent over the shared control interaction time $T$ when using each robot response alternative.}
\label{tab:pr_arm_total_cost_rbr_alt}
\end{table}

Figure \ref{fig:pr_arm_robot_br_states} compares the state $x$ evolution for each of the robot's responses alternatives. The task is achieved by all alternatives, with our proposal showing a smoother behavior while accomplishing it compared to the more aggressive performance of the alternatives using the bank of policies directly. Figure \ref{fig:pr_arm_robot_br_inputs} show the input values for each alternative, which explains the behavior we see in the state plots. Our best response proposal remains at lower values than the alternatives. Although the simulated human input policy $u_h$ is the same for the three scenarios, being a function of the state generates the effect of also having a higher human effort on the inputs when the state starts to increase its values and the agents seek to cooperate to achieve the task. Figure \ref{fig:pr_arm_robot_rbr_only_inputs} zooms in on a shorter time window (from $t=0$ to $t = 90$ seconds) to better show the behavior of the agents' control inputs when the robot is using the closed-form solution of the distribution aware one-step optimization to decide how to act on the system.

Finally, Table \ref{tab:pr_arm_total_cost_rbr_alt} computes the finite-horizon cost $J_{r, T}$ over the shared control interaction period $T$, exhibiting numerically the differences we visualize in the plots before. The lower control inputs and consequent lower state values, together with the faster accomplishment of the task translate to a more optimal performance when using the robot best response.

\begin{remark}
    During the Shared Control interaction no learning process is involved, meaning that our proposal and the alternatives against which is computed leverage the already existing bank of policies. Our PHM estimation process requires the acquisition of the candidate policies through the iterative training procedure, and aims to match the irrationality of human behavior according to the human's performance during the shared control interaction. However, we assume that the task and intention of the human operator are known and that they remain fixed during the interaction. Accounting for the uncertainty in human's intention would require additional computational cost, and potential re-training of the candidate policies. By reducing the computational demand of computing the robot's response, resources can be utilized more efficiently and get closer to meet the requirements for real practical applications.
\end{remark}

\section{Conclusions}

During shared control interactions, human behavior might not always follow the predefined assumptions upon which the robot cooperation has been designed. Adapting the robot control input to observed human behavior presents an opportunity to improve the joint performance by taking into account the irrationality involved in human decision-making. We present the computation of a level-k model of bounded rational behaviors using an iterative training procedure based on ADP, which provides a bank of policies for the corresponding level-k potential human behaviors. We present an estimation process of a probabilistic human model based on the level-k policies to describe the unknown human behavior acting on the system. We define the robot's best response to the human control actions to be the solution of a one-step optimization problem over the complete estimated probabilistic distribution of human behavior based on the level-k candidate policies. In addition, for a specific case of quadratic basis functions and Euler discretization when computing the terminal cost, we derive an efficient closed-form solution to obtain the robot's best response. The effectiveness of the proposed methods is explored through simulations on two different nonlinear systems, showing that the proposed shared control mechanism achieves the desired stabilization task while taking into account human irrational behavior. They demonstrate the capability of the online estimation process to describe human behavior based on the level-k model, and support the decision of defining the robot's response considering the complete probabilistic human model. Future research will focus on extending the proposed mechanism to different tasks in diverse shared-control scenarios, relaxing the limitations on the human intention assumptions, as well as exploring the inclusion of an allocation mechanism accounting for the uncertainty of the PHM estimation process and human performance.

\bmsubsection*{Financial Disclosure}

This work was supported by the Business Finland project AURORA "Automated and Connected Machines" and by the NVIDIA Academic Grant Program through the provision of RTX PRO 6000 Blackwell Max-Q GPUs.

\bmsubsection*{Conflicts of Interest}

The authors declare no conflicts of interest.



\end{document}